\documentclass[journal]{IEEEtran}
\usepackage{lineno,hyperref}
\hypersetup{hidelinks,
	colorlinks=true,
	allcolors=black,
	pdfstartview=Fit,
	breaklinks=true}
\usepackage{bbm}
\usepackage{bm}
\usepackage{amsfonts}
\usepackage{amsmath}
\usepackage{mathrsfs}
\usepackage{amssymb}
\usepackage{enumerate}
\usepackage{graphicx}
\usepackage{subfigure}
\usepackage{booktabs}
\usepackage{graphicx}
\usepackage{setspace}
\usepackage{algorithm}
\usepackage{algorithmic}
\usepackage{setspace}
\usepackage{multirow}
\usepackage{color}
\usepackage{cite}
\usepackage{array}
\usepackage{times}
\usepackage{adjustbox} 
\usepackage{mathptmx}
\usepackage{dsfont}
\usepackage{booktabs} 
\usepackage{tabularx} 

\ifCLASSINFOpdf
\else
\fi
\begin{document}
	\title{Phrase Decoupling Cross-Modal Hierarchical Matching and Progressive Position Correction for Visual Grounding}
	\author{Minghong Xie, Mengzhao Wang, Huafeng Li, Yafei Zhang, Dapeng Tao \IEEEmembership{}, Zhengtao Yu 
		\thanks{This work was supported in part by the National Natural Science Foundation of China under Grant 62276120, and the Yunnan Fundamental Research Projects (202301AV070004, 202401AS070106).}
		\thanks{M. Xie, M. Wang, H. Li, Y. Zhang and Z. Yu are with the Faculty of Information Engineering and Automation, Kunming University of Science and Technology, Kunming 650500, China.(E-mail: minghongxie@163.com (M. Xie) mengzhaowangg@163.com (M. Wang); lhfchina99@kust.edu.cn (H. Li))}
		\thanks{D. Tao is with FIST LAB, School of Information Science and Engineering, Yunnan University, Kunming 650091, China.}
		\thanks{}
\thanks{Manuscript received IEEE Transactions on Multimedia;}}
\markboth{Journal of \LaTeX\ Class Files}%
{Shell \MakeLowercase{\textit{et al.}}}
\maketitle
\begin{abstract}
Visual grounding has attracted wide attention thanks to its broad application in various visual language tasks. Although visual grounding has made significant research progress, existing methods ignore the promotion effect of the association between text and image features at different hierarchies on cross-modal matching. This paper proposes a Phrase Decoupling Cross-Modal Hierarchical Matching and Progressive Position Correction Visual Grounding method. It first generates a mask through decoupled sentence phrases, and a text and image hierarchical matching mechanism is constructed, highlighting the role of association between different hierarchies in cross-modal matching. In addition, a corresponding target object position progressive correction strategy is defined based on the hierarchical matching mechanism to achieve accurate positioning for the target object described in the text. This method can continuously optimize and adjust the bounding box position of the target object as the certainty of the text description of the target object improves. This design explores the association between features at different hierarchies and highlights the role of features related to the target object and its position in target positioning. The proposed method is validated on different datasets through experiments, and its superiority is verified by the performance comparison with the state-of-the-art methods. The source code of the proposed method is available at \href{https://github.com/X7J92/VGNet}{\textcolor{blue}{https://github.com/X7J92/VGNet}}.

\end{abstract}
\IEEEpeerreviewmaketitle
\section{Introduction}
Visual grounding~\cite{deng2023transvg++,du2022visual,yang2022improving,jiao2023suspected,9470913,yang2023improving,9939075} aims to locate the corresponding target object described in the text. Unlike traditional object detection~\cite{wang2021exploring,8740604}, it does not rely on predefined class labels and can detect objects solely based on the content of textual description. Visual grounding plays a guiding role in downstream computer vision tasks, such as image captioning~\cite{li2022comprehending,hu2022scaling,barraco2022unreasonable}, visual question answering~\cite{gupta2022swapmix,ding2022mukea,shao2023prompting}, and image generation~\cite{zhou2022towards,li2023gligen,kim2023dense}. It involves cross-modal matching between textual description and local image features, which requires the grounding model to understand the high-level semantic features of text and images deeply, and to correlate the features associated with the two modalities implicitly. Due to the modality gap between text and image, it is challenging to achieve these goals. Current research in visual grounding primarily focuses on supervised settings, and its performance on the standard benchmark still far from expectations, which implies that the visual grounding remains challenging even in supervised settings.
\begin{figure}[t!]
\centering

\includegraphics[width=0.9\linewidth]{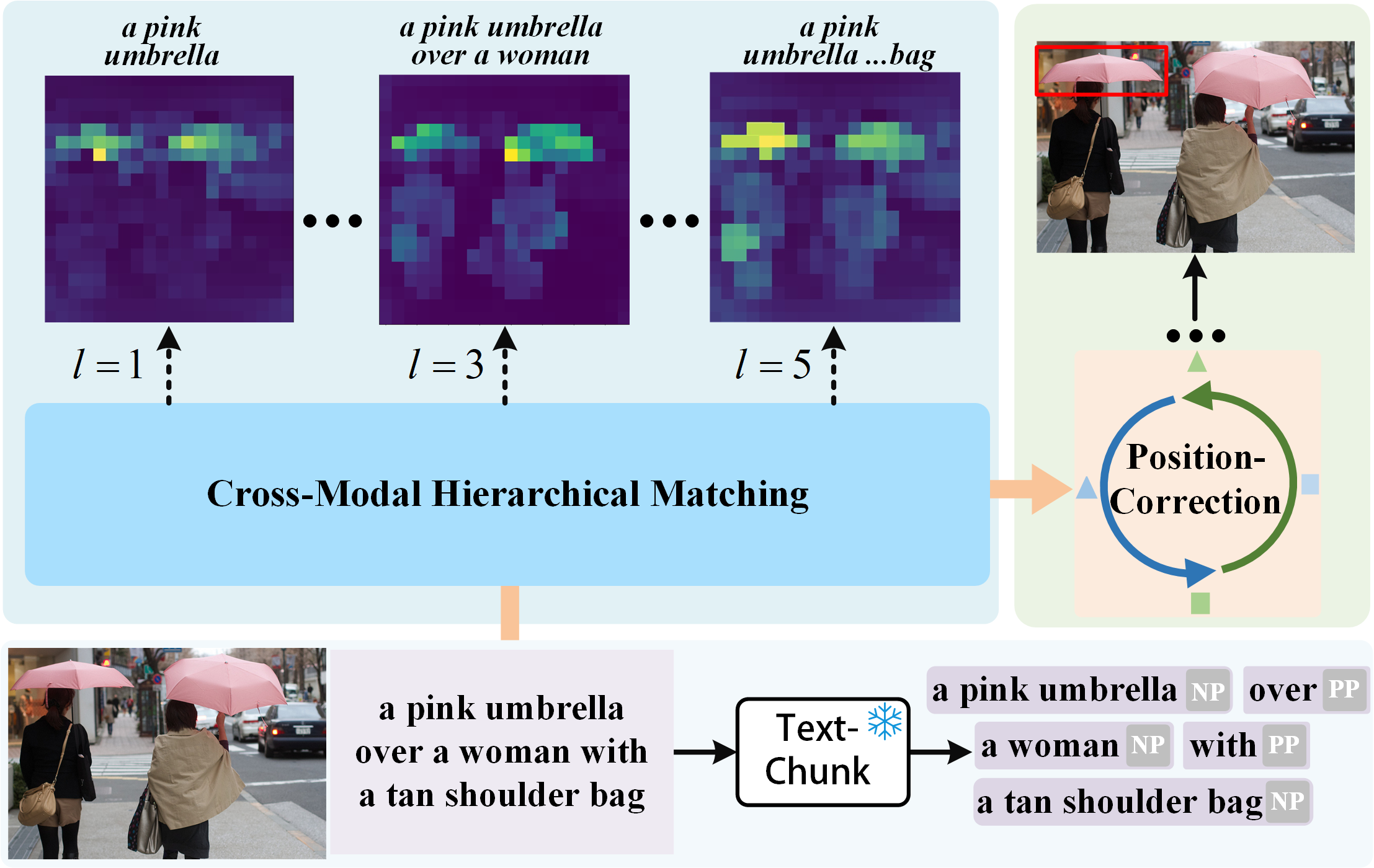}
\caption{Main idea of the proposed method. As shown in this figure, the shared objects described in different levels of text are likely to be the target objects. Therefore, the target area will be clearer if shared features are highlighted.}
\label{Fig1}
\end{figure}

This work is conducted in a supervised setting and primarily explores the construction of cross-modal matching mechanisms between textual description and local image region. Based on whether the region proposals of the objects are used to assist in target localization, existing supervised visual grounding methods can be categorized into two types: two-stage visual grounding~\cite{Alpher05,8691415,9009000,chen2021ref} and one-stage visual grounding~\cite{9806393,9699024,10035452,9982652,9710016}. The two-stage methods rely on pre-trained object detectors~\cite{wang2021exploring,8740604} to obtain proposals for all objects in the image. They search for the object described in the text by measuring the similarity between the text features and the features of all proposals. These methods benefit from the assistance of object detectors, making cross-modal matching more convenient. However, this type of method is highly dependent on the performance of the object detector. The failure of the object detector to identify objects may result in false negatives during cross-modal matching. On the contrary, the one-stage methods can avoid this issue. Due to the lack of assistance from object detectors, one-stage methods need to address the problem arising from missing region proposals. The most popular way is to fuse visual features with text features and directly use the fused features for object localization \cite{huang2021look,qiu2020language,ye2021one,9699024,9710016}. These methods boost feature representation by fusing the textual and visual features, enabling visual grounding even without the assistance of region proposals. However, they only consider the correspondence between local or global text and image features while neglecting to explore the role of the association between text and image features at different hierarchies in target localization. 

Fig.~\ref{Fig1} illustrates the text ``A pink umbrella over a woman with a tan shoulder bag" and its corresponding image. One can observe that the phrase ``a pink umbrella" only refers to one pink umbrella. Still, in the image, there are two objects that match this phrase. However, when the description is extended to ``a pink umbrella over a woman," it implies a scene in which the umbrella is positioned above someone. The corresponding image content becomes increasingly specific as the description becomes more refined. Each hierarchical text phrase points to a shared target object. As the hierarchy increases, the position of the text-indicated target in the image is gradually highlighted, while non-specified objects remain stable or even suppressed. It means that when different hierarchical phrases interact with visual features from the previous hierarchy, the features corresponding to the described object will gradually receive more attention, as seen in Fig.~\ref{Fig1} with the pink umbrella and the pedestrian's shoulder bag. If this progressive mechanism can be leveraged to emphasize the role of features in the target object's corresponding region, it would improve detection performance.

This paper proposes the Phrase Decoupling Cross-Modal Hierarchical Matching and Progressive Position Correction for visual grounding. The method belongs to the one-stage visual grounding approach but differs from most existing methods because it does not rely on feature fusion to achieve object localization. To achieve progressive hierarchies relationship exploration among features, a hierarchical mask generation mechanism is designed. This mechanism can separate features at different hierarchies, facilitating the relationship exploration between hierarchies. Then, a hierarchical matching scheme for text and image is constructed to emphasize the role of keywords in cross-modal matching. Furthermore, with the hierarchical matching mechanism, a corresponding progressive position correction method is proposed, which can progressively correct the target object's position based on the matching results between hierarchies. This progressive strategy continuously refines and adjusts the location of the target object's bounding box as the certainty of the target object described in the text improves. This design explores the association between features at different hierarchical levels and highlights the features corresponding to the target object at different hierarchies. The main contribution of this work can be summarized as follows.
\begin{itemize}
\item We propose a progressive hierarchical association mining approach, which establishes a structured hierarchical association between text and image features, highlighting the effect of features aligned with the target object.
\item  Based on the hierarchical matching association between text and image, we further design a progressive position correction scheme for target object detection, achieving precise localization of the target object specified in the text.
\item Experimental results on three referring expression datasets demonstrated this approach's effectiveness and its superiority compared to the state-of-the-art methods. Additionally, it also exhibited a certain advantage in terms of computational efficiency.
\end{itemize}

\begin{figure*}[t!]
\centering
\includegraphics[width=0.99\linewidth]{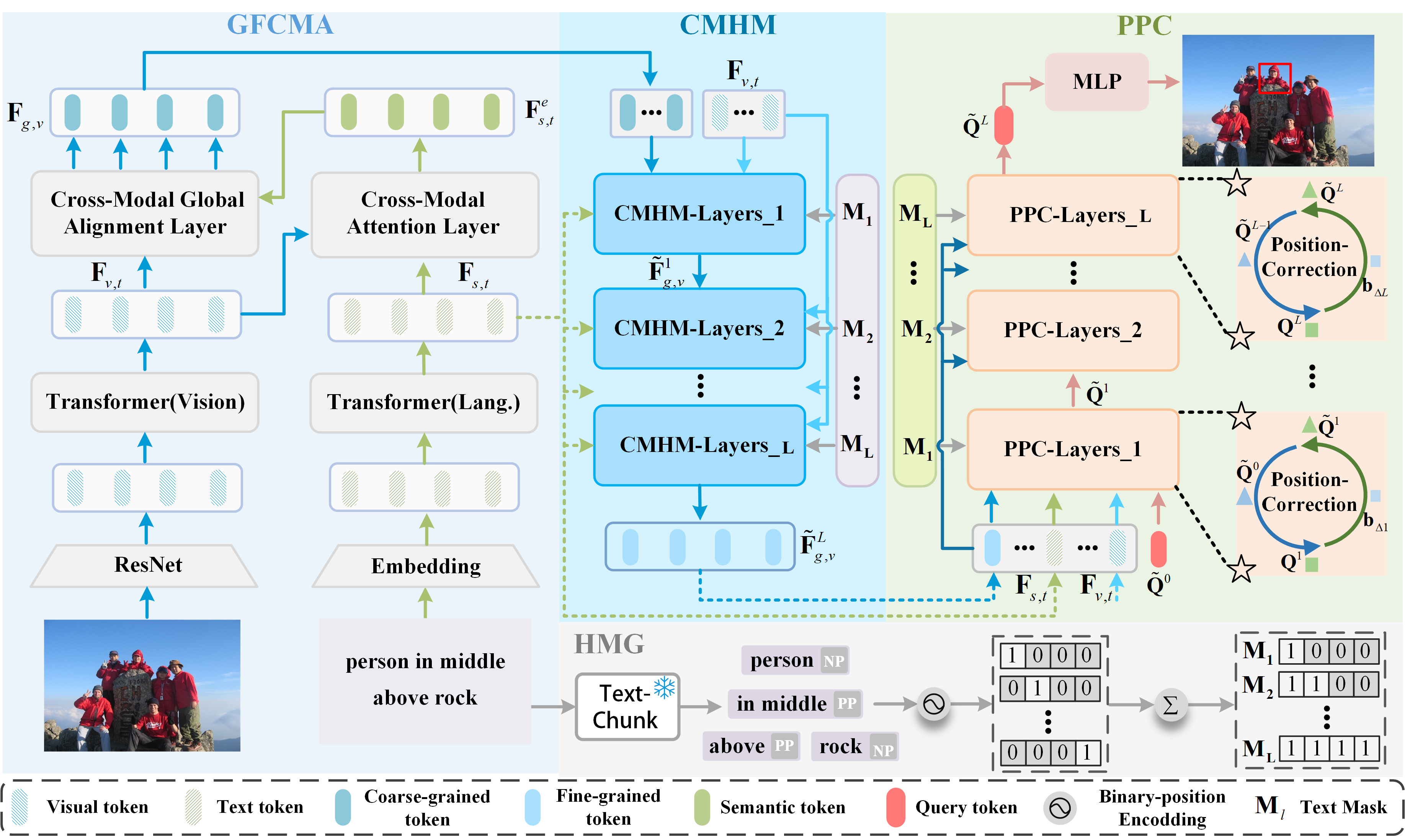}
\caption{Architecture of the proposed method. Given the input image and the text, GFCMA extracts features across different modalities and establishes a global relationship between text and image. Meanwhile, HMG parses phrases from the input text and uses them to create masks. Then, these features and masks are sent into CMHM to produce more discriminative features for the referred object. Finally, PPC utilizes the results of hierarchical matching to progressively correct the target object's position, achieving localization of the target object.}
\label{Fig2}
\end{figure*}

\section{Related Work}
\subsection{Two-stage Visual Grounding}
Two-stage visual grounding first uses a pre-trained object detection model to generate region proposals and then selects the proposals that best match the textual description. In the first stage, detectors such as Fast R-CNN~\cite{girshick2015fast} or Mask R-CNN~\cite{he2017mask} are commonly used. In the second stage, proposals are matched with the provided textual description, and the proposal with the highest matching score is selected as the prediction output. To ensure accurate matching, a typical approach is to model the relationships between proposals and language descriptions by introducing a graph structure and transforming the visual grounding task into an optimal matching problem of graph nodes~\cite{jing2020visual,yang2019dynamic,yang2019cross,8999516}. However, such methods do not consider the impact of syntactic structures on text features, which limits the expressive power of text features. Parsing and understanding the semantics of sentences using syntactic tree structures and then matching them with proposals can alleviate this issue~\cite{8691415,9009000}. Moreover, MattNet~\cite{Alpher05} employs a modular design strategy to describe the object in the text, and its position and relationship with the subject. To enhance the cross-modal feature matching capability, CM-Att-Erase~\cite{liu2019improving} generates challenging samples by erasing words from the text or image content, emphasizing the remaining information's role in cross-modal matching.  Two-stage methods leverage object detection models to locate described objects, mitigating the challenges of blind matching between text and target regions. However, those methods require high precision of the object detection results. When the object detector fails or generates inaccurate region proposals, it may directly impact the subsequent matching of text and targe features.

\subsection{One-stage Visual Grounding}
The one-stage visual grounding typically involves 1) fusing visual and textual features, 2) partitioning the fused features into grids, and 3) performing position regression for each grid. This approach eliminates the reliance on object detectors in cross-modal matching. For example, FAOSA-VG~\cite{yang2019fast} concatenates visual and textual features and feeds them into YOLOv3~\cite{redmon2018yolov3} to regress the bounding boxes of the target objects. ReSC~\cite{yang2020improving} splits the textual description using attention mechanism and performs multiple rounds of inference between the image and text to gradually reduce referring ambiguity, addressing the limitations of previous one-stage methods when understanding complex queries. RCCF~\cite{liao2020real} treats visual grounding as a correlation filtering process, mapping text features from the language domain to the visual domain and using the mapping results as templates (kernels) to filter image features to generate the center positions and sizes of target boxes. 

Recently, the Transformer architecture has been extensively studied. It was initially introduced in \cite{vaswani2017attention} for neural machine translation. Influenced by the successful application of the self-attention mechanism, researchers have further explored applying the Transformer framework to the vision tasks, such as image classification~\cite{chen2021crossvit,dosovitskiy2020image},action recognition~\cite{10066256,xu2023pyramid,9954217,9782720}, and visual grounding~\cite{9710016,ye2022shifting,zhan2023rsvg,10298801,9806393,10269126}, among other tasks. In visual grounding tasks, transformers are used to enhance the interaction between vision and text, thereby improving visual grounding performance. Specifically, TransVG~\cite{9710016} uses a series of simple transformer encoders for multimodal fusion and inference in visual grounding, avoiding the need for complex handcrafted fusion modules. Dynamic MDETR~\cite{10298801} decouples the whole grounding process into encoding and decoding stages and uses sparse priors on object locations to develop a dynamic multimodal transformer decoder to improve the efficiency of the visual grounding process. CLIP-VG~\cite{10269126} employs adaptive curriculum learning and a simple end-to-end network to achieve visual grounding based on CLIP~\cite{radford2021learning}. Word2Pix~\cite{9806393} utilizes a one-stage visual grounding network with an encoder-decoder transformer structure to achieve attention learning from words to pixels. Although these methods are effective, they do not match the text with the described objects at different hierarchies, failing to highlight the association between text and image. This limits further improvement in matching performance. The proposed method establishes a connection between text and image features across hierarchies within a one-stage framework, in which a progressive refinement of the target regions can be achieved.

\section{The Proposed Method}
\subsection{Overview}
The visual grounding model proposed in this paper comprises four components: Global Feature Cross-Modal Alignment (GFCMA), Hierarchical Mask Generation (HMG), Cross-Modal Hierarchical Matching (CMHM), and Progressive Position Correction (PPC), as shown in Fig.~\ref{Fig2}. GFCMA primarily focuses on establishing global relationship between text and image, providing global information for the subsequent creation of hierarchical association. HMG parses phrases from the input text and uses them to create masks for CMHM. Driven by the hierarchical mask, CMHM achieves cross-modal feature hierarchical matching, allowing features to be associated across different hierarchies. PPC uses the hierarchical matching results to progressively correct the target object position. In the following sections, we will elaborate on each component.

\subsection{Global Feature Cross-Modal Alignmehnt}
GFCMA mainly comprises a visual encoding branch and a textual encoding branch (Fig.~\ref{Fig2}). Assuming that the source image fed into the visual encoding branch is $v$, its corresponding sentence is \( s \). \( v \) is first input into ResNet101 to extract shallow features $\hat{\mathbf{F}}_v \in \mathbb{R}^{C \times H \times W}$. $C$, $H$, and $W$ denote the number of channels, height, and width of $\hat{\mathbf{F}}_v$, respectively. For each pixel position of $\hat{\mathbf{F}}_v$, the feature map is vectorized to obtain the feature $\mathbf{F}_v=\left[\mathbf{f}_v^1, \mathbf{f}_v^2, \cdots, \mathbf{f}_v^{N_v}\right]$, composed of tokens $\mathbf{f}_v^i \in \mathbb{R}^{C \times 1}(i=1,2 \cdots, N_v)$, where $N_v=H\times W$. $\mathbf{F}_v$ is input into the Transformer layer, resulting in $\mathbf{F}_{v, t}=\left[\mathbf{f}_{v, t}^1, \mathbf{f}_{v, t}^2, \cdots, \mathbf{f}_{v, t}^{N_v}\right]$, where
$\mathbf{f}_{v, t}^i \in \mathbb{R}^{C \times 1}(i=1,2 \cdots, {N_v})$. For textual branches, assume that the features extracted from text via the Transformer layer are $\mathbf{F}_{s, t}=\left[\mathbf{f}_{s, t}^1, \mathbf{f}_{s, t}^2, \cdots, \mathbf{f}_{s, t}^{N_s}\right]$, composed of tokens $\mathbf{f}_{s, t}^j \in \mathbb{R}^{C \times 1}(j=1,2 \cdots, N_s)$, where $N_s$ is the number of textual tokens.  

\begin{figure}[t!]
\centering
\includegraphics[width=\linewidth]{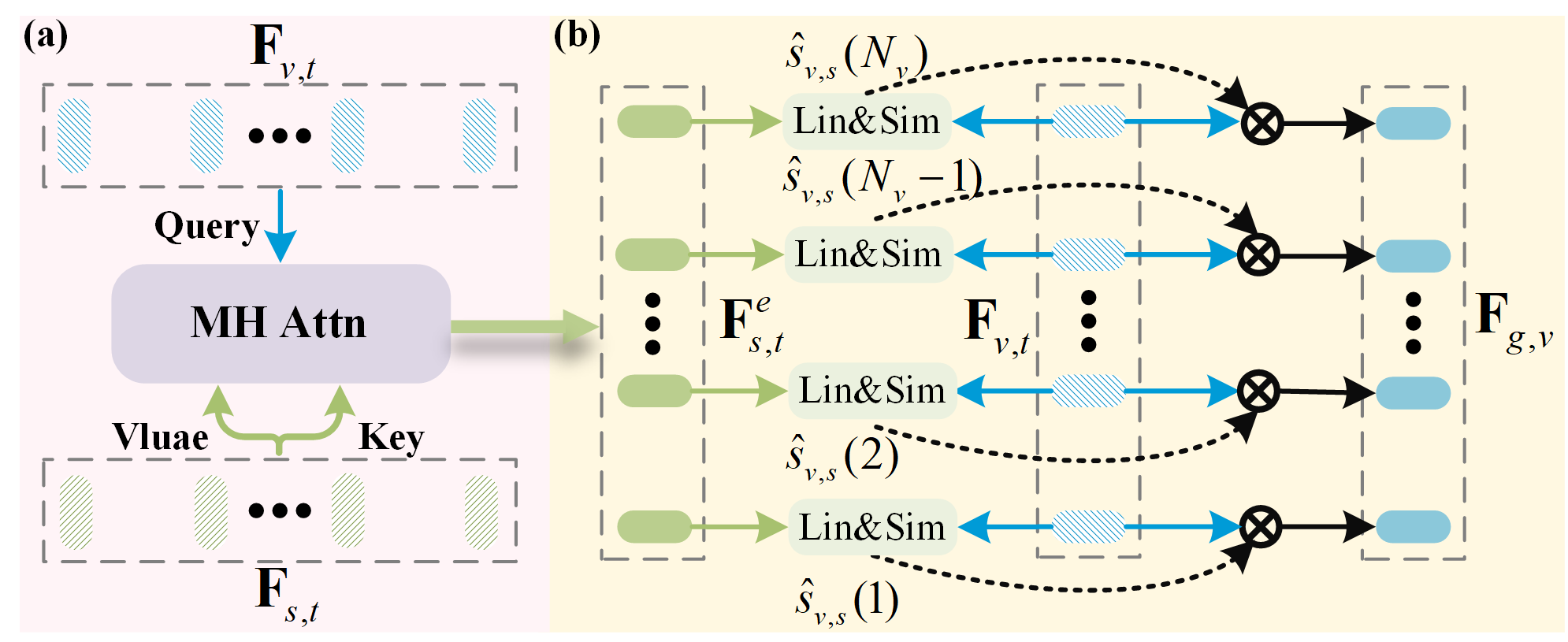}
\caption{Details of the cross-modal attention layer and the cross-modal global alignment layer. (a) Cross-modal attention layer. (b) Cross-modal global alignment layer. Lin\&Sim denotes the linear mapping and similarity measure.}
\label{Fig3}
\end{figure}
As shown in Fig.~\ref{Fig3}, $\mathbf{F}_{v, t}$ and $\mathbf{F}_{s, t}$ pass through a cross-modal attention layer and a cross-modal global alignment layer to obtain aligned coarse-grained features. $\mathbf{F}_{v, t}$ is utilized to highlight the corresponding features between $\mathbf{F}_{s, t}$ and $\mathbf{F}_{v, t}$. Specifically, in the cross-modal attention layer, $\mathbf{F}_{v, t}$ is transformed into the query vector $\mathbf{Q}_{v, t}$ via linear mapping, while $\mathbf{F}_{s, t}$ is linearly mapped to both the key vector $\mathbf{K}_{s, t}$ and the value vector $\mathbf{V}_{s, t}$. The semantic feature $\mathbf{F}_{s, t}^e$ is then obtained through multi-head attention (MH Attn), with each token in $\mathbf{F}_{s, t}^e$ incorporating global information from the text.

Furthermore, in the cross-modal global alignment layer, the visual feature $\mathbf{F}_{v, t}$ and the semantic feature $\mathbf{F}_{s, t}^e$ are linearly mapped into the same semantic space. The resulting mapped features are denoted as $\mathbf{F}_{v, t}^{\prime}$ and $\mathbf{F}_{s, t}^{\prime e}$, respectively. Then, we can establish a global relationship between the tokens in $\mathbf{F}_{s, t}^{\prime e}$ and the corresponding tokens in $\mathbf{F}_{v, t}^{\prime}$, highlighting the feature tokens of the targeted object described in the text, without the need for pairwise feature metrics. Specifically, the similarity between the corresponding tokens of $\mathbf{F}_{v, t}^{\prime}$ and $\mathbf{F}_{s, t}^{\prime e}$ can be measured as:
\begin{equation} 
s_{v,s}(i)=\frac{\mathbf{F}_{v, t}^{\prime T}(:, i) \mathbf{F}_{s, t}^{\prime e}(:, i)}{\|\mathbf{F}_{v, t}^{\prime}(:, i)\|_2 \times\|\mathbf{F}_{s, t}^{\prime e}(:, i)\|_2}
\end{equation}
By using the following formula, $s_{v,s}(i)$ is normalized to $[0,1]$, resulting in \( \hat{s}_{v,s} \):
\begin{equation}  
\hat{s}_{v,s}(i)=\frac{\exp \left(\lambda^* s_{v,s}(i)\right)}{\sum_{i=1}^{N_v} \exp \left(\lambda^* s_{v,s}(i)\right)}
\end{equation}
where $\lambda^*$ is the inverse temperature of the softmax function. Each value \( \hat{s}_{v,s} \) indicates the degree of relevance between the visual token and the corresponding textual token. Leveraging  \( \hat{s}_{v,s} \), we obtain the globally aligned cross-modal features:
\begin{equation}
\mathbf{F}_{g, v}=\left[\hat{s}_{v,s}(1) \mathbf{f}_{v, t}^1, \hat{s}_{v,s}(2) \mathbf{f}_{v, t}^2, \cdots, \hat{s}_{v,s}(N_v) \mathbf{f}_{v, t}^{N_v}\right]
\end{equation}
The cross-modal global alignment layer emphasizes the related features between images and text, suppressing the non-related ones. This provides support for subsequent hierarchical feature matching.

\subsection{Hierarchical Mask Generation}
There is a hierarchical correspondence between the text phrase and the potential target area. That is, the more phrases describing the target object, the more accurate the target object position in the image. With this hierarchical association, we devise a feature cross-modal hierarchical matching scheme. In detail, we propose a hierarchical mask generation mechanism to achieve hierarchical matching of features. For texts that require phrase decoupling, we employ the phrase decoupling model from~\cite{akbik2018coling} to extract related noun phrases, verb phrases, and adjective phrases, etc. Take the text ``white and black cat laying on orange cat" as an example. After model decoupling of this text, and we can obtain four phrases: ``white and black cat" (noun phrase), ``laying" (verb phrase), ``on" (prepositional phrase), and ``orange cat" (noun phrase). We assign a specific binary positional encoding to each phrase $\mathbf{P}=\left\{\mathbf{p}_1, \mathbf{p}_2, \ldots, \mathbf{p}_L\right\}$, where $\mathbf{p}_i = [\mathbf{0}, \ldots, \underset{i}{\mathbf{1}}, \ldots, \mathbf{0}] \quad (i = 1,2, \ldots, L)$ denotes the position encoding of the $i$-th phrase, $\mathbf{0} \in \mathbb{R}^{1 \times N_i}$ is a vector entirely constituted of 0,  $\mathbf{1} \in \mathbb{R}^{1 \times N_i}$ is a vector fully made of 1 and $N_i$ is the number of words that make up the current phrase. The mask at the $j$-th hierarchy is:
\begin{equation} 
\mathbf{M}_j=\sum_{i=1}^j \mathbf{p}_i
\end{equation}
\subsection{Cross-Modal Hierarchical Matching}

\begin{figure}[t!] 
\centering
\includegraphics[width=\linewidth]{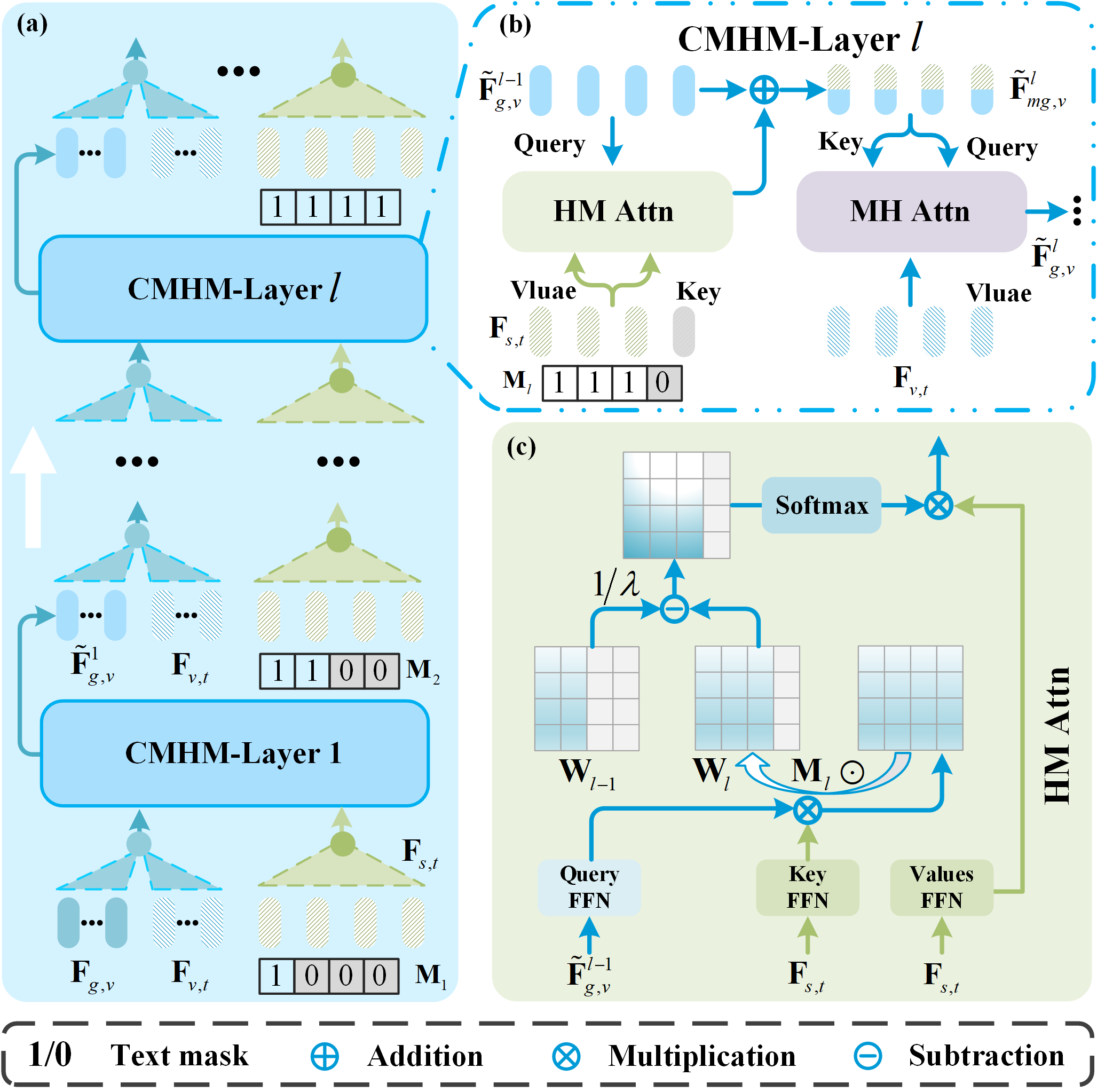}
\caption{Details of the cross-modal hierarchical matching (CMHM). (a) Structure of the CMHM . (b) Structure of CMHM-Layer. (c) Structure of hierarchical mask attention (HM Attn).}

\label{Fig4}
\end{figure}

In the paper, we present CMHM for cross-modal matching, which relies on the generated hierarchical mask, as shown in Fig.~\ref{Fig4}. CMHM consists of multiple CMHM layers, each including a Hierarchical Mask Attention (HM Attn) and a Multi-head Attention (MH Attn). HM Attn is used to extract features consistent with hierarchical masks, while MH Attn highlights cross-modal consistent features. The number of hierarchies formed by the textual phrases corresponds to the number of layers in CMHM. For the $l$-th CMHM layer, the hierarchical mask can be represented as $\bm{M}_l$, which consists of $[\mathbf{1}, \cdots, \underset{l}{\mathbf{1}}, \mathbf{0}, \cdots, \mathbf{0}]$. Assuming the input features to this layer are $\tilde{\mathbf{F}}_{g, v}^{l-1}$, $\mathbf{F}_{v, t}$ and $\mathbf{F}_{s, t}$, where $\tilde{\mathbf{F}}_{g, v}^{l-1}$ is the output of the $(l-1)$-th CMHM. When $l=1$, $\tilde{\mathbf{F}}_{g, v}^{0}=\mathbf{F}_{g, v}$.
By performing linear mapping, $\tilde{\mathbf{F}}_{g, v}^{l-1}$ is transformed into the query vector $\mathbf{Q}_{h, v}^l$, and $\mathbf{F}_{s, t}$ is mapped to both the key vector $\mathbf{K}_{h, t}^l$ and the value vector $\mathbf{V}_{h, t}^l$. Thus, the resulting output by HM Attn can be represented as:

\begin{small}
\begin{equation}
	\mathbf{F}_m^l=\operatorname{softmax}\left(\mathbf{W}_{l} -\frac{1}{\lambda} \mathbf{W}_{l-1}\right) \left(\mathbf{V}_{h, t}^l\right)^T
\end{equation}
\end{small}
where $\mathbf{W}_{l}=\frac{(\mathbf{Q}_{h, v}^l)^T\mathbf{K}_{h, t}^l}{\sqrt{d_{2}}} \odot \mathbf{M}_l$, $d_{2}$ indicates the dimensionality of $\mathbf{K}_{h, t}^l$, $\odot$ denotes the element-wise multiplication. $\lambda$ is a modulating factor greater than $1$. It is used to retain the visual information corresponding to the current hierarchical text query while appropriately weakening the visual information corresponding to the previous hierarchical text query. This ensures that the model does not overly rely on the previous hierarchical information during hierarchical cross-modal matching. Instead, it can fully utilize the new information of the current hierarchy in addition to referencing the previous hierarchical information.

To enhance the consistent information between $\mathbf{F}_m^l$ and $\tilde{\mathbf{F}}_{g, v}^{l-1}$, they are summed to produce:
\begin{equation}
\tilde{\mathbf{F}}_{m g, v}^l=\mathbf{F}_m^l+ \tilde{\mathbf{F}}_{g, v}^{l-1}
\end{equation}
After different linear mapping, the results of $\tilde{\mathbf{F}}_{m g, v}^l$ serve as Query and Key, and the visual feature $\mathbf{F}_{v, t}$ serves as Value in MH Attn, yielding an enhanced feature $\tilde{\mathbf{F}}_{ g, v}^l$. Guided by the hierarchical text masks, we achieve a fine-grained interaction between textual and visual features. This interaction allows $\tilde{\mathbf{F}}_{ g, v}^l$ to incorporate hierarchical textual information consistently. This addresses the limitations of coarse-grained matching and leverages the association across different hierarchies to boost cross-modal matching. It is worth noting that the context between words is retained by segmenting the text at the phrase level, allowing the model to understand these relationships and further encourage its fine-grained matching more comprehensively.

\subsection{Progressive Position Correction}

\begin{figure}[t!]
\centering
\includegraphics[width=\linewidth]{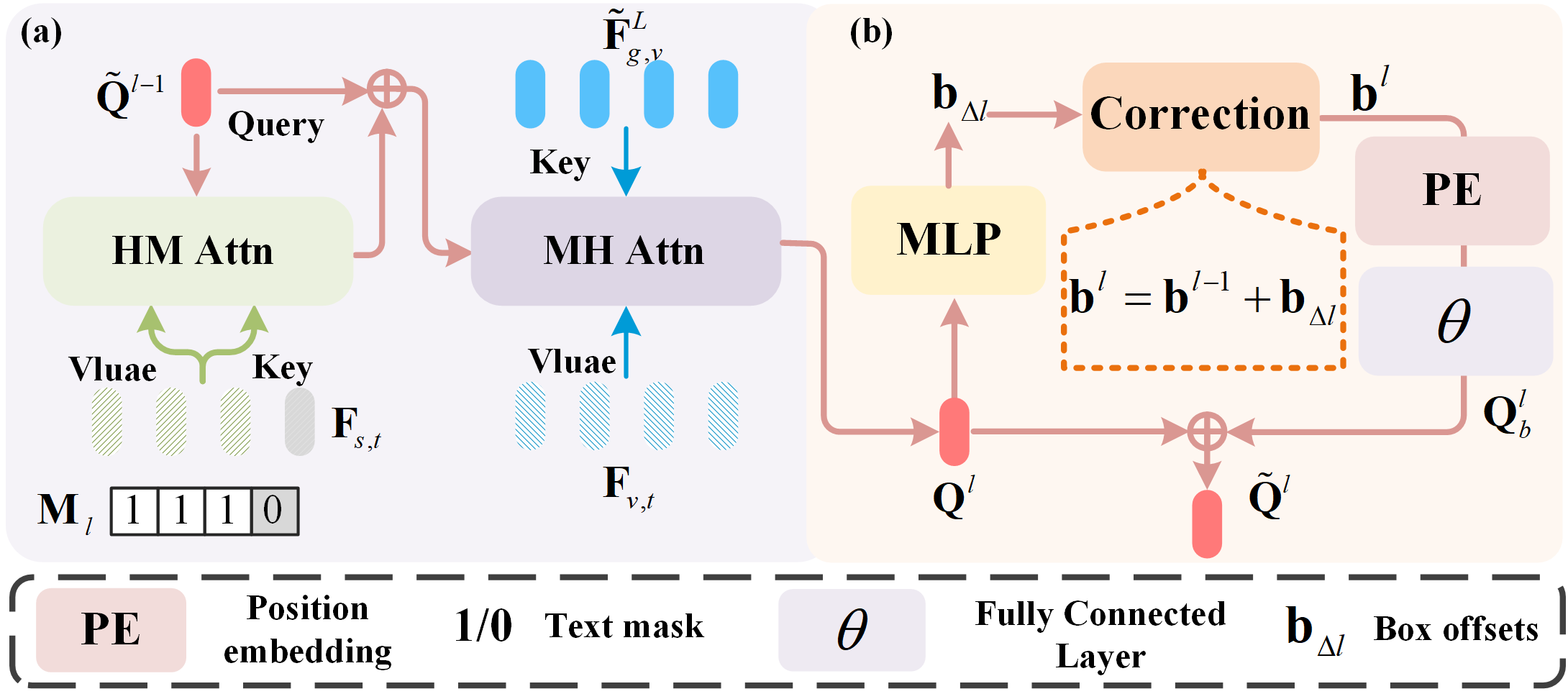}
\caption{Illustration of progressive position correction (PPC) layer. (a) Hierarchical semantic aggregation (HSA). (b) Hierarchical position correction (HPC).}
\label{Fig5}
\end{figure}

With the guidance of hierarchical matching, we develop a novel PPC module for target localization, which achieves precise localization by continuously refining the target's position. As illustrated in Fig.~\ref{Fig2}, the PPC module consists of $L$ PPC layers. The PPC layer comprises Hierarchical Semantic Aggregation (HSA) and  Hierarchical Position Correction (HPC), as depicted in Fig.~\ref{Fig5}. HSA is used to aggregate both textual and visual feature at the current hierarchy, consisting of HM Attn and MH Attn. HM Attn is used to extract features consistent with hierarchical masks, while MH Attn is used to highlight cross-modal consistent features. HPC is used to refine the position information of the target object from the previous layer by utilizing the target's location data at the current layer, thereby strengthening the connections across layers.

Assume that the inputs of the $l$-th PPC layer are the target query $\tilde{\mathbf{Q}}^{l-1}$, textual features $\mathbf{F}_{s, t}$, hierarchical masks $\mathbf{M}_l$, visual features $\mathbf{F}_{v, t}$, and $\tilde{\mathbf{F}}_{g, v}^L$, and the output is $\tilde{\mathbf{Q}}^l$. In the first PPC layer, a random vector is used as $\tilde{\mathbf{Q}}^{0}$. In the $l$-th PPC layer, the HSA's output is denoted as $\mathbf{Q}^l$. 
Since $\mathbf{Q}^l$ incorporates both textual and visual feature from the current layer, it carries more information about the target object compared to $\tilde{\mathbf{Q}}^{l-1}$. These information are directly related to the positional information of the target object. Consequently, $\mathbf{Q}^l$ can be used to refine the predicted position information from the previous hierarchy. Let $\mathbf b^{l-1}$ be the bounding box position predicted in the $(l-1)$-th hierarchy, $\mathbf b^l$ (when $l=0$, $\mathbf b^0$ is a position vector consisting of 0) be the corrected bounding box position, then the change $\mathbf b_{\Delta l}$ from $\mathbf b^{l-1}$  to $\mathbf b_l$ can be predicted by:
\begin{equation} 
\mathbf b_{\Delta l}=\sigma\left(\operatorname{MLP}\left(\mathbf{Q}^l\right)\right)
\end{equation}	
where $\sigma$ represents the sigmoid activation function, and MLP represents a multi-layer perceptron. In this case, the bounding box position predicted in the current stage can be expressed as:
\begin{equation}
\mathbf b^l=\mathbf b^{l-1}+\mathbf b_{\Delta l}
\end{equation}

In HPC, the current position information plays a crucial role in predicting the position at the next hierarchy. Therefore, we embed the position information predicted in the current stage into $\mathbf{Q}^l$ to assist the position predicting of the next hierarchy. Suppose the coordinate of $\mathbf b^l$ is $\left[x_l, y_l, w_l, h_l\right]$, where $\left(x_l, y_l\right)$ denotes the center point coordinate of the bounding box, $w_l$ and $h_l$ respectively denote the height and width of the box. Since the dimensions of $\mathbf{b}^l \in \mathbb{R}^{1 \times 4}$ and $\mathbf{Q}^l \in \mathbb{R}^{1 \times C}$ are not consistent, we adopt the method of~\cite{kenton2019bert} to encode the position of $\mathbf{b}^l$, then project it to the same dimension with $\mathbf{Q}^l$, resulting in $\mathbf{Q}_b^l \in \mathbb{R}^{1 \times C}$:
\begin{equation}
\mathbf{Q}_b^l=\theta\left(\left[\operatorname{PE}\left(x_l\right) ; \operatorname{PE}\left(y_l\right) ; \operatorname{PE}\left(w_l\right) ; \operatorname{PE}\left(h_l\right)\right]\right)
\end{equation}	

\begin{table*}[htbp]
\caption{Comparison (Top-1 \textit{Prec@0.5}) with state-of-the-art methods on the RefCOCO, RefCOCO+, RefCOCOg, and ReferItGame datasets. `Val` is the validation set. `Test` is the testing set. VLTVG* shows our replicated results. The bolded data represents the optimal result.}\label{tab1}	
\renewcommand{\arraystretch}{1} 
\begin{adjustbox}{width=1\linewidth,keepaspectratio}
	\begin{tabular}{c|c|c c c| c c c |c c|c|c}
		\hline
		\hline
		\multirow{2}{*}{ Models } & \multirow{2}{*}{ Backbone } & \multicolumn{3}{c}{ RefCOCO } & \multicolumn{3}{|c|}{$\mathrm{RefCOCO}+$} & \multicolumn{2}{c|}{ RefCOCOg } & \multicolumn{1}{c|}{ ReferItGame } & \multicolumn{1}{c}{Time}  \\
		
		& & val ($\%$) & testA ($\%$) & testB ($\%$) & val ($\%$) & testA ($\%$) & testB ($\%$) & val ($\%$) & test ($\%$) & test ($\%$) & ($ms$) \\
		
		\hline
		\multicolumn{12}{l}{ \textbf{Two-stage} } \\
		
		\hline
		
		CMN~\cite{8099953} & Vgg16 & - & 71.03 & 65.77 & - & 54.32 & 47.76 & - & - & - & -  \\

		VC~\cite{8753569} & Vgg16 & - & 73.33 & 67.44 & - & 58.40 & 53.18 &- & - & - & -  \\
		
		ParalAttn~\cite{8578545} & Vgg16 & - & 75.31 & 65.52 & - & 61.34 & 50.58 & - & - & - & -  \\

		DGA~\cite{yang2019dynamic} & Vgg16 & - & 78.42 & 65.53 & - & 69.07 & 51.99 & - & - & 63.28 & -  \\

		RvG-Tree~\cite{8691415} & ResNet-101 & 75.06 & 78.61 & 69.85 & 63.51 & 67.45 & 56.66 & 66.95 & 66.51 & - & - \\
		
		NMTree~\cite{9009000} & ResNet-101 & 76.41 & $81.21$ & 70.09 & $66.46$ & $72.02$ & 57.52 & 65.87 & 66.44 & - & - \\
		
		Ref-NMS~\cite{chen2021ref} & ResNet-101 & 78.82 & 82.71 & 73.94 & 66.95 & 71.29 & 58.40 & 68.89 & 68.67 & - & - \\
		
		CM-Att-Erase~\cite{liu2019improving} & ResNet-101 & 78.85 & 83.14 & 71.32 & 68.09 & 73.65 & 58.03 & 67.99 & 68.07 & - & - \\
		
		MAttNet~\cite{Alpher05} & ResNet-101 & 76.65 & 81.14 & 69.99 & 65.33 & 71.62 & 56.02 & 66.58 & 67.27 & 29.04 & 385  \\
		\midrule 
		\multicolumn{12}{l}{ \textbf{One-stage} } \\
		\midrule

		RealGIN~\cite{9470913} & DarkNet-53 & 77.25  & 78.70 & 72.10 & 62.78 & 67.17 & 54.21 & - & 62.75 & 62.33 & - \\
		
		FAOA~\cite{yang2019fast} & DarkNet-53 & 71.15 & 74.88 & 66.32 & 56.86 & 61.89 & 49.46 & 59.44 & 58.90 & 59.30 & - \\

		MCN~\cite{9157418} & DarkNet-53 & 80.08 & 82.29 & 74.98 & 67.16 & 72.86 & 57.31 & 66.46 & 66.01 & - & - \\
		
		ReSC~\cite{yang2020improving} & DarkNet-53 & 77.63 & 80.45 & 72.30 & 63.59 & 68.36 & 56.81 & 67.30 & 67.20 & 64.60 & - \\
		
		LBYL~\cite{9578259} & DarkNet-53 & 79.67 & 82.91 & 74.15 & 68.64 & 73.38 & 59.49 & - & - & 67.47 & - \\
		
		SeqTR~\cite{zhu2022seqtr} & DarkNet-53 & 81.23 & 85.00 & 76.08 & 68.82 & 75.37 & 58.78 & 71.35 & 71.58 & 69.66 & - \\
		
		TRAR~\cite{9711011} & DarkNet-53 & - & 81.40 & 78.60 & - & 69.10 & 56.10 & 68.90 & 68.30 & - & - \\

		CTMDI~\cite{wu2024improving} & DarkNet-53 & 82.59 & 85.13 & 79.82 & 69.52 & 75.53 & 63.48 & 73.43 & 73.05 & - & - \\

		PFOS~\cite{9699024} & ResNet-101 & 78.44 & 81.94 & 73.61 & 65.86 & 72.43 & 55.26 & 67.89 & 67.63 & 67.90 & - \\

		PLV-FPN~\cite{liao2022progressive} & ResNet-101 & 81.93 & 84.99 & 76.25 & 71.20 & 77.40 & 61.08 & 70.45 & 71.08 & - & - \\
		
		Word2Pixel~\cite{9806393} & ResNet-101 & 81.20 & 84.39 & 78.12 & 69.71 & 76.11 & 61.24 & 70.81 & 71.34 & - & 56 \\

		TransVG~\cite{9710016} & ResNet-101 & 81.02 & 82.72 & 78.35 & 64.82 & 70.70 & 56.94 & 68.67 & 67.73 & 70.73 & 40 \\

		VLTVG*~\cite{yang2022improving} & ResNet-101 & 83.10 & 85.43 & 78.65 & 71.52 & 76.06 & 61.28 & 72.63 & 71.52 & 68.76 & \textbf{30}\\

		$\mathbf{Ours}$ &ResNet-101 & $\mathbf{83.81}$ & $\mathbf{85.56}$ & $\mathbf{79.95}$ & $\mathbf{73.56}$ & $\mathbf{77.41}$ & $\mathbf{64.80}$ & $\mathbf{74.75}$ &$ \mathbf{73.50}$ & $\mathbf{71.40}$ & $32$ \\
		\hline

		D-MDETR~\cite{10298801} & ResNet-50 & 82.80 & 84.99 & 77.74 & 69.05 & 74.20 & 60.17 & 70.73 & 70.68 & 69.48 & - \\
		
		$\mathbf{Ours}$ & ResNet-50 & $\mathbf{83.13}$ & $ \mathbf{85.02} $ & $\mathbf{78.24}$ & $\mathbf{71.22}$ & $\mathbf{76.59}$ & $\mathbf{62.35}$ & $\mathbf{73.80}$ &$ \mathbf{72.39}$ & $\mathbf{70.55}$ & $\mathbf{29}$ \\\hline
		
	    LUNA~\cite{liang2023luna} & Swin-Base & 86.25 & 88.31 & \textbf{82.73} & 76.01 & 80.72 & \textbf{68.46} & 76.91 & 76.65 & 73.89 & - \\
		
		$\mathbf{Ours}$ &Swin-Base & $\mathbf{86.72}$ & $\mathbf{88.52}$ & $82.04$ & $\mathbf{77.89}$ & $\mathbf{82.23}$ & $68.39$ & $\mathbf{77.83}$ &$ \mathbf{77.21}$ & $\mathbf{75.37}$ & $\mathbf{37}$ \\\hline

		TransVG++~\cite{deng2023transvg++} & ViT-Base & 86.28 & 88.37 & 80.97 & 75.39 & 80.45 & $\mathbf{66.28}$ & 73.86 & 76.30  & 74.70 & - \\
		
	    JMRI~\cite{10285487} & ViT-Base & 82.97 & 87.30 & 74.62 & 71.17 & 79.82 & 57.01 & 71.96 & 72.04 & 68.23 & - \\

		$\mathbf{Ours}$ &ViT-Base & $\mathbf{86.87}$ & $\mathbf{88.73}$ & $\mathbf{81.53}$ & $\mathbf{76.33}$ & $\mathbf{81.52}$ & 65.89 & $\mathbf{76.62}$ &$ \mathbf{76.74}$ & $\mathbf{75.41}$ & $\mathbf{34}$ \\
		\hline

		CLIP-VG~\cite{10269126} & CLIP & 84.29 & 87.76 & 78.43 & 69.55 & 77.33 & 57.62 & 73.18 & 72.54 & 70.89 & 47 \\

		$\mathbf{Ours}$ &CLIP & $\mathbf{85.10}$ & $\mathbf{87.98}$ & $\mathbf{81.20}$ & $\mathbf{74.28}$ & $\mathbf{78.54}$ & $\mathbf{63.02}$ & $\mathbf{76.51}$ &$ \mathbf{75.48}$ & $\mathbf{73.86}$ & $\mathbf{36}$ \\
		
		\hline
		\hline
	\end{tabular}
\end{adjustbox}
\end{table*}

where PE is the sine positional encoding function, $[.]$ represents concatenation along the channel dimension, and $\theta$ represents a fully connected layer. By adding $\mathbf{Q}_b^l$ to  $\mathbf{Q}^l$, we obtain the target query for the $l+1$ hierarchy:
\begin{equation}
\tilde{\mathbf{Q}}^l=\mathbf{Q}_b^l+\mathbf{Q}^l
\end{equation}

We perform $N$ iterations on the PPC module. The output $\tilde{\mathbf{Q}}_n^L$ of the $L$-th PPC-layer at $n$-th iteration is used to predict the bounding box $B_n$ corresponding to the textual reference. $B_n$ can be generated as:
\begin{equation}
B_n=\operatorname{MLP}\left(\tilde{\mathbf{Q}}_n^L\right)
\end{equation}	
where MLP represents a multi-layer perceptron. We use the Generalized IoU loss ($l_{g}$)~\cite{8953982} and $l_{1}$-loss to constrain $B_n$ in each iteration. The details are as follows:
\begin{equation}
\mathcal{L}_Q=\sum_{n=1}^N \lambda_1 \ell_1\left(B_{g t}, B_n\right)+\lambda_2 \ell_{\mathrm{g}}\left(B_{g t}, B_n\right)
\end{equation}	
where $B_{gt}$ represents the ground truth, $\lambda_1$ and $\lambda_2$ are balancing weights and $N$ is the total number of iterations. Let $\mathbf{b}_{n}^{L}$ be the position predicted by Eq.(8) at the $L$-th hierarchy and $n$-th iterations. Additionally, we use the $\ell_1$ to make $B_{gt}$ and $\mathbf{b}_{n}^{L}$ consistent:
\begin{equation}
\mathcal{L}_{\text {cons }}=\sum_{n=1}^N \ell_1\left(B_{g t}, \mathbf{b}_{n}^{L}\right)
\end{equation}	
The entire network is trained in an end-to-end manner, and model parameters are optimized using the total loss $\mathcal{L}_{\text {total }}$:
\begin{equation}
\mathcal{L}_{\text {total }}=\mathcal{L}_{\text {cons }}+\mathcal{L}_Q
\end{equation}

\section{Experiments}
\subsection{Datasets and Evaluation Protocol}
\textbf{Datasets}. The proposed method is evaluated on visual grounding datasets, including RefCOCO~\cite{1111111}, RefCOCO+~\cite{1111111}, RefCOCOg~\cite{7780378} and ReferItGame~\cite{kazemzadeh2014referitgame}.  Those datasets consists of images and their corresponding referring expressions.

\textbf{RefCOCO/RefCOCO+/RefCOCOg} The RefCOCO~\cite{1111111}, RefCOCO+~\cite{1111111} and RefCOCOg~\cite{7780378} datasets are created by selecting images and target objects from MSCOCO~\cite{vinyals2016show}. The RefCOCO dataset consists of $19,994$ images, which contains $142,210$ referring expressions for $50,000$ target objects. The RefCOCO+ dataset contains $19,992$ images with $141,564$ referring expressions for 49,856 objects. RefCOCOg includes $25,799$ images and $95,010$ expressions for $49,822$ objects. For the RefCOCO and RefCOCO+ datasets, they are divided into four sections: train, val, testA, and testB. Among them, testA focuses on images containing multiple persons, while testB involves images with a variety of non-human objects. The images in the RefCOCO and RefCOCO+ datasets contain multiple objects of the same category. Moreover, in RefCOCO+, the referring expressions exclude the words that denote absolute position (such as 'left'). The RefCOCOg dataset is divided into train, val, and test sections, with its images typically containing $2$ to $4$ objects of the same category. In addition, the average text length of RefCOCOg is $8.43$ words, which is significantly longer than that of the RefCOCO and RefCOCO+ datasets ($3.61$ and $3.53$ words, respectively)

\textbf{ReferItGame} The ReferItGame dataset~\cite{kazemzadeh2014referitgame} is created by selecting images and target objects from SAIAPR12~\cite{escalante2010segmented}. It contains $120,072$ expressions for $19,987$ objects within $20,000$ natural scene images. The ReferItGame is divided into train, val, and test sets. 

\textbf{Evaluation Protocol}. We follow the method used in the work of \cite{9710016} to evaluate the detection performance. A prediction is deemed correct and denoted as \textit{Prec@0.5} when the Intersection over Union (IoU) between the predicted bounding box and its Ground Truth surpasses the threshold of $0.5$.

\subsection{Implementation Details}
In this paper, the maximum text length is set to $40$ words and the input images are resized to $640\times 640$. For visual encoding branch, we use ResNet-50/101~\cite{he2016deep} and $6$ transformer layers to extract visual features. For textual encoding branch, the basic BERT-base-uncased model~\cite{kenton2019bert} is used for initialization. We employ the phrase decoupling model from~\cite{akbik2018coling} for phrase decoupling. During training, we use the AdamW~\cite{loshchilov2018decoupled} optimizer with an initial learning rate of $10^{-4}$ and batch size of $64$ for model training. We set $\lambda_{1}=5$ and $\lambda_{2}=1$ in the loss function described in Eq.(13). To ensure a fair comparison with more methods, we added ViT-Base\cite{dosovitskiy2020image} and Swin-Base\cite{liu2021swin} as visual backbones, while keeping the text encoder unchanged. Inspired by the large-scale pre-training visual language models, we use the visual-language model CLIP~\cite{radford2021learning} as the backbone of the proposed method. To improve training stability, we fix the CLIP encoder and introduce an MLP layer for feature modulation. All other settings are consistent with the original configurations. The training is conducted on four RTX4090 GPUs for $90$ epochs.

\subsection{Comparison with State-of-the-Art Methods}
The experimental results of different methods on the ReferItGame, RefCOCO, RefCOCO+ and RefCOCOg datasets are compared in Table~\ref{tab1}. Those comparative methods are divided into two-stage and one-stage methods. The proposed method belongs to the one-stage methods. Compared to the one-stage methods using ResNet-50 or ResNet-101 as the backbone, the proposed method outperforms all the comparison methods. Specifically, when we use ResNet-50 as the visual backbone, the proposed method achieves a $3.07\%$ performance improvement on the RefCOCOg validation dataset compared to the D-MDETR\cite{10298801}. Additionally, when using ResNet-101, the proposed method achieves a $3.52\%$ increase on the RefCOCO+ testB dataset compared to the VLTVG\cite{yang2022improving}. When we replace the ResNet-101 with ViT-base, the proposed method achieves a $8.88\%$ improvement compared to JMRI\cite{10285487} on the RefCOCO+ testB dataset. Additionally, when we use Swin-Base as the visual backbone, the proposed method achieves a $3.25\%$ improvement compared to LUNA\cite{liang2023luna} on the RefCOCOg validation dataset. Furthermore, we compare the proposed method with CLIP-VG which is developed based on a large-scale visual language model. For a fair comparison, we replace the backbone in the proposed method with CLIP. As shown in Table~\ref{tab1}, the proposed method outperforms the CLIP-VG, demonstrating a significant enhancement of $4.73\%$ on the RefCOCO+ validation dataset. Additionally, it exhibits advancements of $2.97\%$ on the ReferItGame test dataset and $5.4\%$ on the RefCOCO+ testB dataset, respectively. It can be attributed to the effectiveness of the proposed hierarchical relationship mining and matching. Moreover, the proposed model using ResNet-50 as its backbone achieves the fastest speed, with an inference time of $29ms$ when evaluated under equivalent conditions.

\begin{figure*}[htbp]
	\centering
	\includegraphics[width=0.90\linewidth]{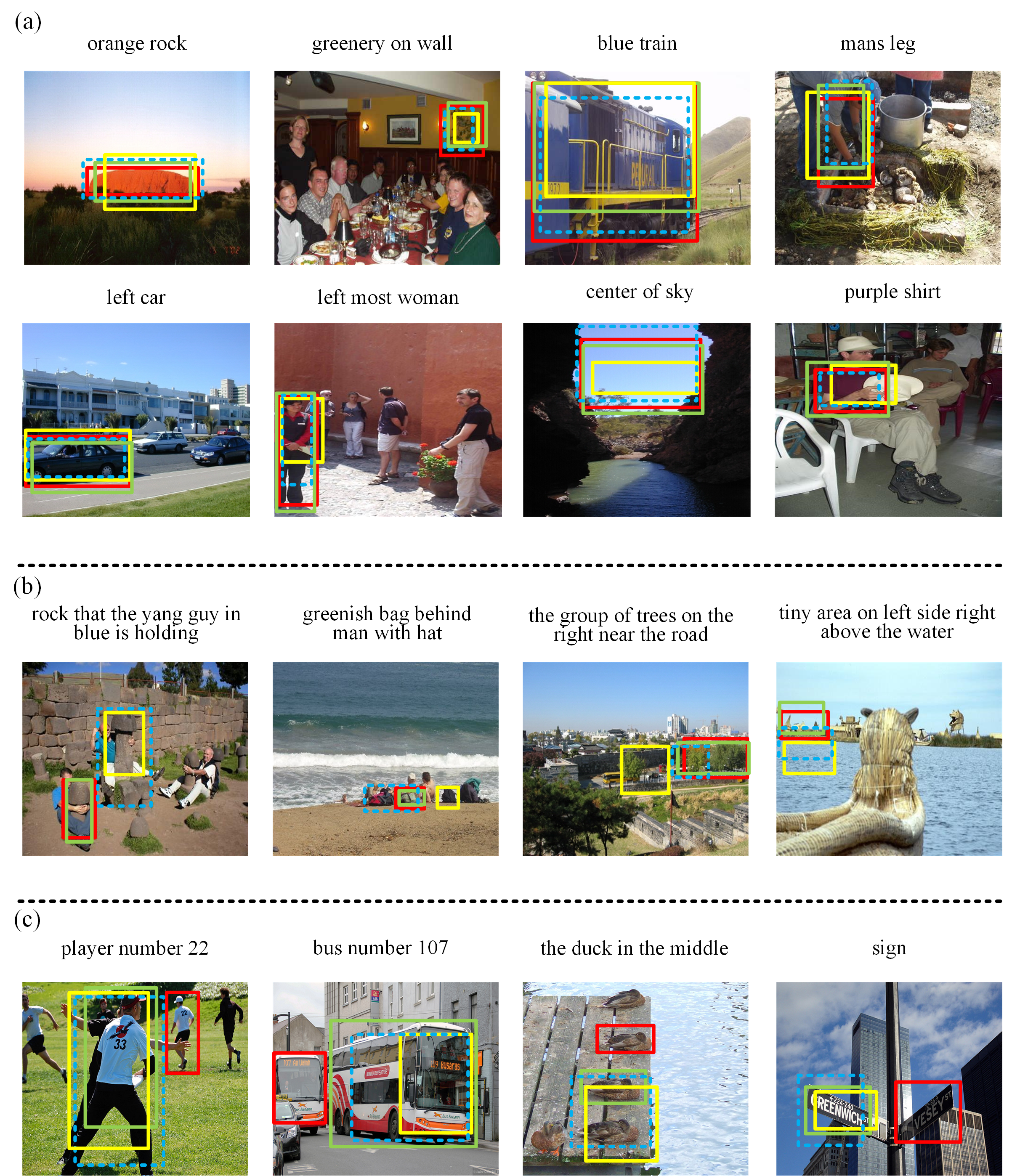}
	\caption{Results predicted by the proposed method , CLIP-VG and Baseline. (a) shows successful predictions by the proposed method, CLIP-VG, and the Baseline when using simple text queries. (b) shows that CLIP-VG and the Baseline fail, but our method succeeds when using complex text queries. (c) shows failed predictions by the proposed method, CLIP-VG, and the Baseline.The red boxes represents the ground truths. The green and yellow boxes denote the prediction results of the proposed method and Baseline, respectively. The blue dashed boxes denotes the prediction results of CLIP-VG.}
	\label{Fig6}
\end{figure*}

In Fig.~\ref{Fig6}, we compare the results predicted by the proposed method, CLIP-VG~\cite{10269126} and Baseline. The red box represents the ground truth. The green and yellow boxes denote the results predicted by the proposed method and Baseline, respectively. The blue dashed box denotes the prediction results of CLIP-VG. Fig.~\ref{Fig6}(a) shows the prediction results when the query images and simple referring expressions are fed to the models. It can be seen that all the methods can focus well on the target objects when the hierarchy of the referring expressions is less. However, as the hierarchy of the referring expressions increases, a significant discrepancy emerges between the labels and the prediction results of both the Baseline and CLIP-VG, as shown in Fig.~\ref{Fig6}(b). In contrast, the proposed method consistently sustains high localization precision. It is attributed to our method's ability to mine the hierarchical relations of text, thereby facilitating precise localization of the target objects. Additionally, Fig.~\ref{Fig6}(c) shows some failure cases for our method, CLIP-VG and the Baseline. In the first two test cases of Fig.~\ref{Fig6}(c), we observe that our model and other methods struggle to accurately locate text queries containing numbers. This difficulty arises from the models' challenges in recognizing numerical information. In the last two failed test cases of Fig.~\ref{Fig6}(c), the ambiguous text queries lead to our method, CLIP-VG and the Baseline predicting boxes that diverge from the ground truth annotations.

\begin{table}[htbp]
\caption{Ablation study of each component on the ReferItGame datasets.}
\label{table2}
\scriptsize 
\begin{tabular}{ccccccc}
	\hline
	\hline
	Baseline & GFCMA & CMHM & PCC w/o HPC & PCC & Val(\%) & Test(\%) \\
	\hline
	\checkmark & & & & & 65.23 & 63.89 \\
	\checkmark & \checkmark & & & & 68.19 & 66.48 \\
	\checkmark & \checkmark & \checkmark & & & 71.33 & 68.54 \\
	\checkmark & \checkmark & \checkmark & \checkmark & & 72.94 & 70.02 \\
	\checkmark & \checkmark & \checkmark &  & \checkmark & 74.15 & 71.40 \\
	\hline
	\hline
\end{tabular}
\end{table}

\begin{figure}[t!]
\centering
\includegraphics[width=\linewidth]{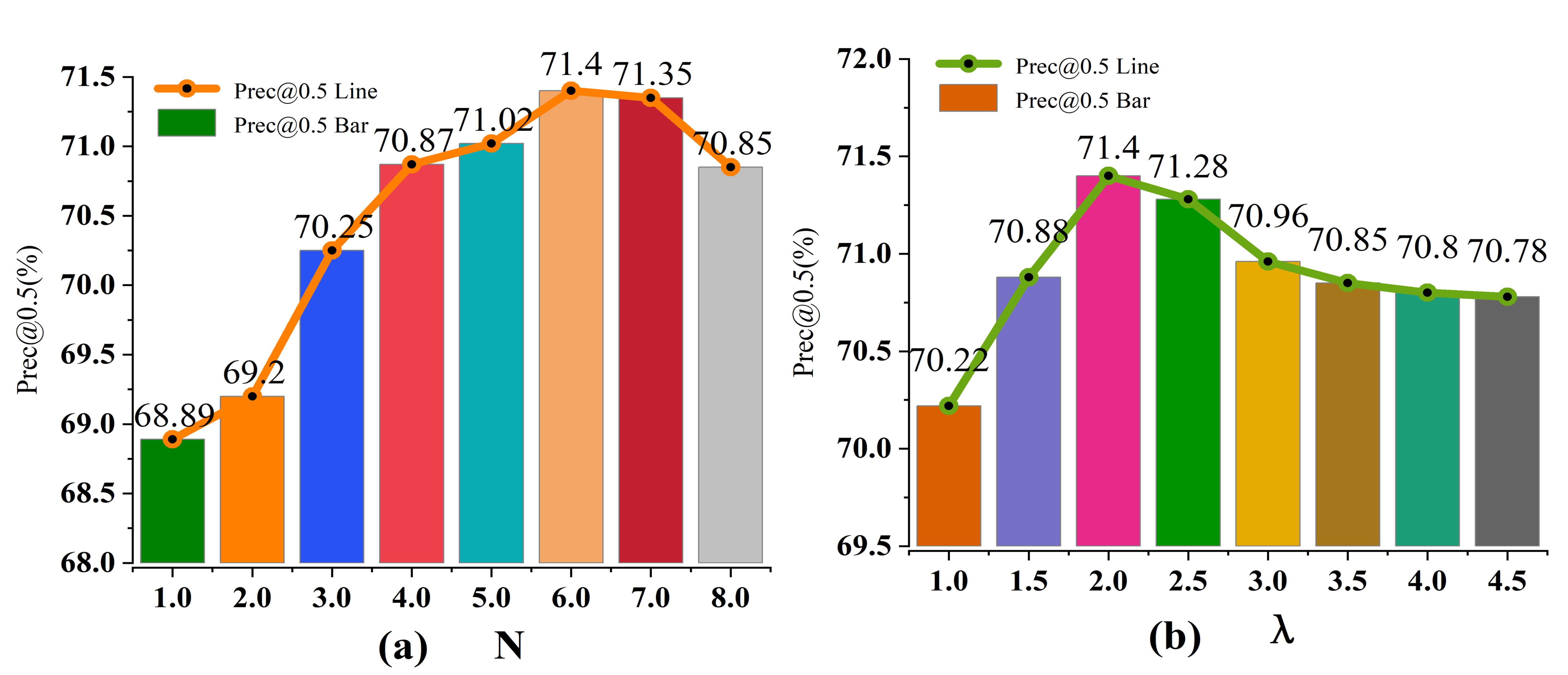}
\caption{Influence of the iteration times $N$ and $\lambda$ on model performance.}
\label{Fig7}
\end{figure}
\begin{figure}[htbp]
\centering
\includegraphics[width=\linewidth]{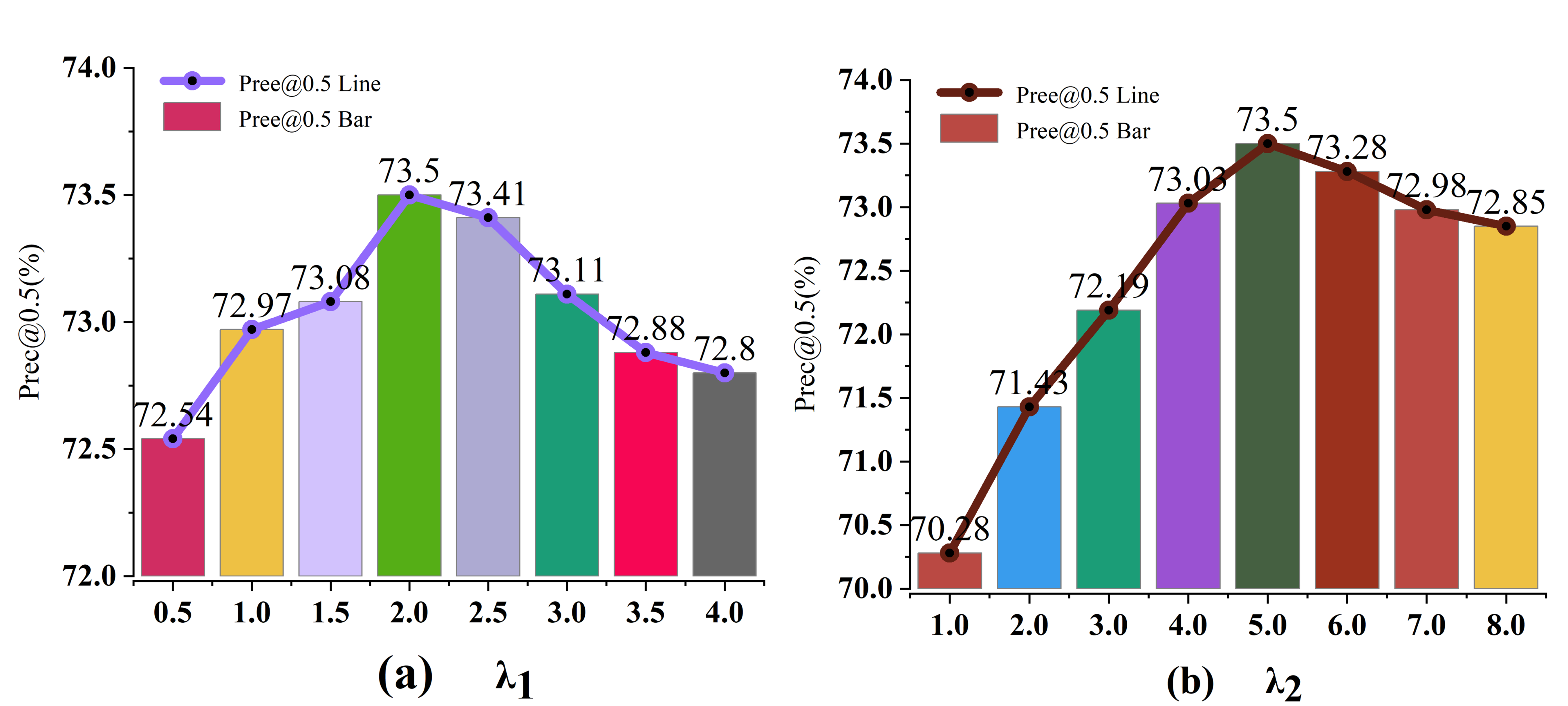}
\caption{Influence of $\lambda_{1}$ and $\lambda_{2}$ on model performance.}
\label{Fig8}
\end{figure}

\subsection{Ablation Study}
The proposed network mainly includes GFCMA, CMHM and PPC modules. We conduct the ablation study on ReferItGame dataset to evaluation their contribution. We remove the cross-modal global alignment layer from the GFCMA module, and the residual part of the GFCMA module is used as the Baseline model. This network is trained with $l_{1}$ loss and generalized IoU loss. The results of ablation experiments are presented in Table~\ref{table2}.

\textbf{Effectiveness of GFCMA.} As shown in Table~\ref{table2}, when the GFCMA is added into the Baseline, the accuracy on \textit{Prec@0.5} is improved by $2.96\%$ and $2.59\%$ on validation and test sets, respectively. The accuracy increases due to that the GFCMA focuses on global relationships between text and image, providing global information for subsequent fine-grained interactions.

\textbf{Effectiveness of CMHM.} When CMHM is added into Baseline+GFCMA, the performance of Baseline+GFCMA+CMHM is further improved on validation and test sets. It implies that the CMHM can mine correlation relationships between different hierarchies in cross-modal matching, and highlight the role of the shared features at different hierarchies.

\textbf{Effectiveness of PCC.} Compared to Baseline+GFCMA+CMHM, the inclusion of PCC results in an increase of $2.82\%$ ($2.86\%$) on the validation (test) set. It is because that the position of the detected bounding box can be gradually corrected by the PCC.

\textbf{Effectiveness of HPC.} To validate the contribution of the HPC in PPC module to the proposed model, the position correction in Eqs.(8)-(11) is removed and $\mathbf{Q}^l$ ($l=1,2,\cdots,L$) aggregates textual and visual information layer by layer. The $\mathbf{Q}^L$ is used to regress and determine the final bounding box location. The model without hierarchical position correction is denoted as ‘PPC w/o HPC’. As observed in Table~\ref{table2}, the addition of HPC results in an increase of $1.21\%$ and $1.38\%$ in $\textit{Prec@0.5}$ on the validation and test sets of ReferItGame, respectively. It is proved that the position correction information $\mathbf{Q}_b^l$ is useful for precise localization of target object.

\begin{figure}[htbp]
	\centering
	\includegraphics[width=0.75\linewidth]{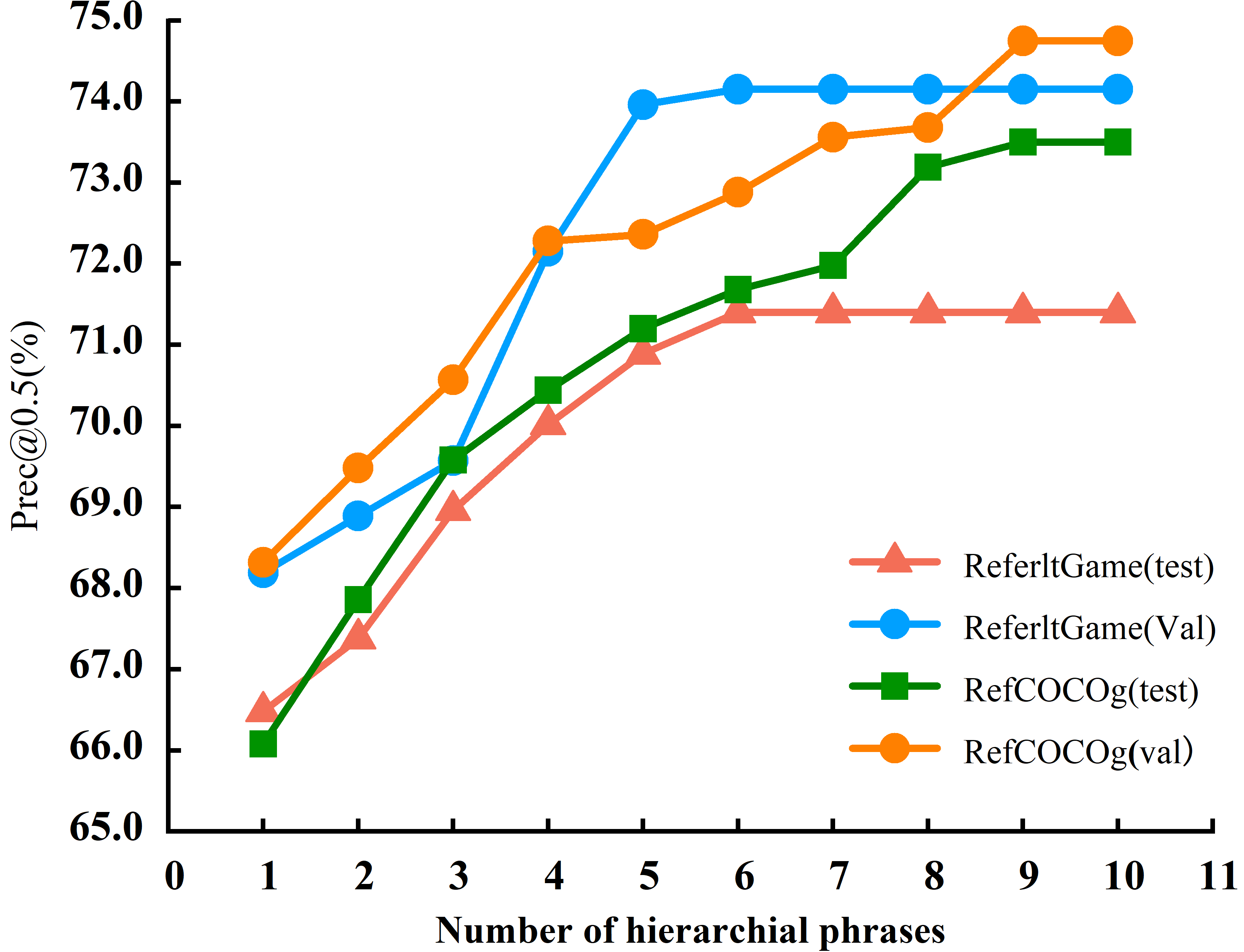} 
	\caption{Influence of the number of hierarchies on model performance.}
	\label{Fig9}
\end{figure}

\subsection{Parameter Analysis} 
\textbf{Analysis of iteration times $N$.} In Fig.~\ref{Fig7}(a), we evaluate influence of the iteration times of PPC. As shown in Fig.~\ref{Fig7} (a), the proposed mehtod achieves the best performance when the iteration times $N$ reaches $6$. Therefore, we set $N=6$ throughout the experiment.

\textbf{Analysis of hyperparameter $\lambda$.} In the proposed method, the hyperparameters $\lambda$ used in Eq. (6) is a constant greater than $1$. To determine the value of $\lambda$, Fig.~\ref{Fig7} (b) shows the changes of model performance when $\lambda$ takes different values. From those results, we can see that the index \textit{Prec@0.5} reaches its maximum when $\lambda=2$. 
\begin{figure*}[htbp]
	\centering
	\includegraphics[width=0.9\linewidth]{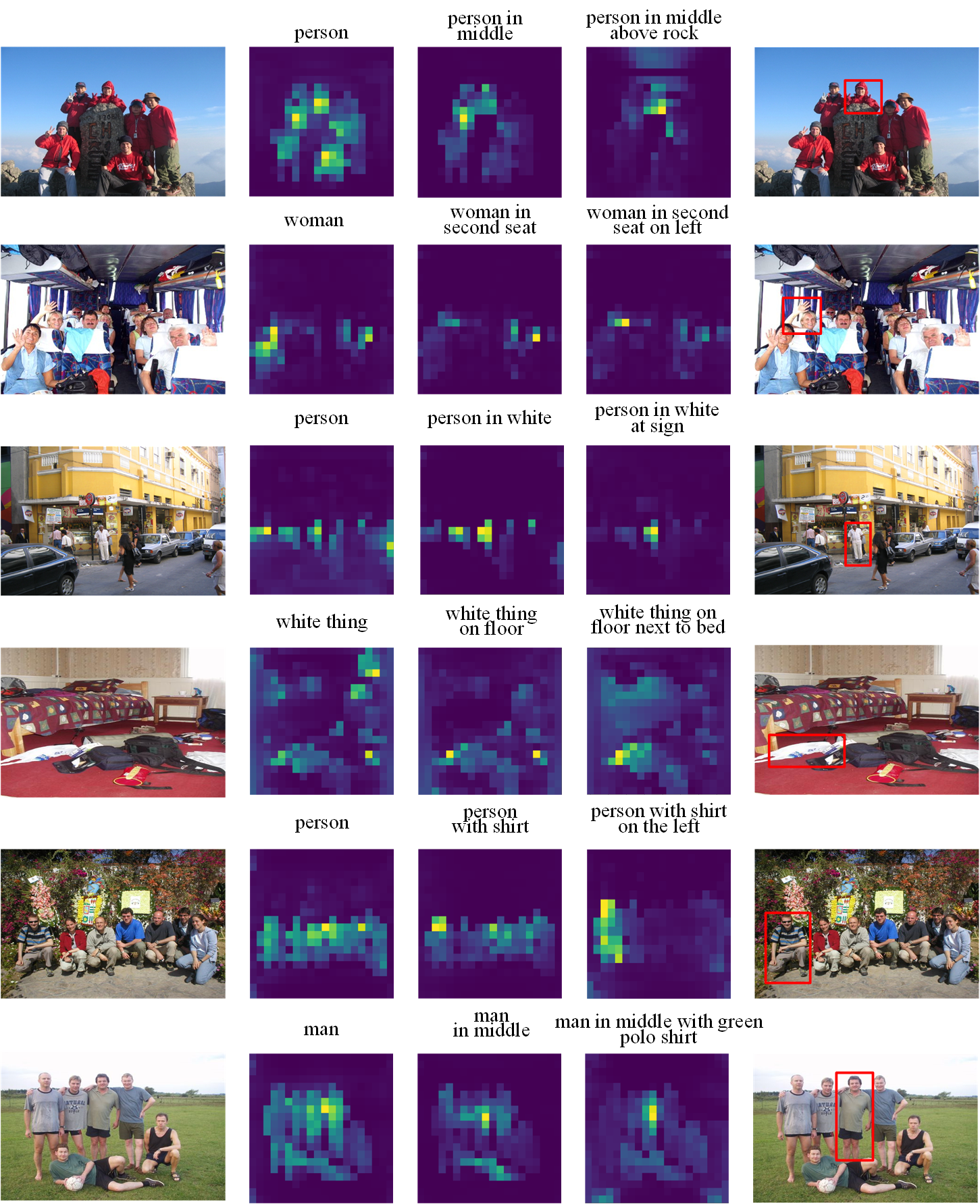}
	\caption{Visualization of attention maps at different hierarchies and the model detection results. In the attention maps shown in the $2^{nd}$, $3^{rd}$ and $4^{th}$ columns, brighter pixels indicate that the image receives more attention from the feature extraction network.}
	\label{Fig10}
\end{figure*}

\textbf{Analysis of hyperparameter $\lambda_{1}$ and $\lambda_{2}$.} In this section, we analyze the impact of the hyperparameters $\lambda_1$ and $\lambda_2$ in Eq.(13) on detection performance on the test set of the RefCOCOg. Fig.~\ref{Fig8} shows the changes of model performance when $\lambda_1$ and $\lambda_2$ take different values, respectively. It is observed from Fig.~\ref{Fig8}(a) that when $\lambda_1=2$, the $\textit{Prec@0.5}$ reaches its peak. Therefore, we set $\lambda_1$ to $2$. Moreover, the proposed mehtod achieves the best performance when $\lambda_2=5$, as shown in Fig.~\ref{Fig8}(b). Consequently, we set $\lambda_2=5$.
\begin{figure}[htbp]
	\centering
	\includegraphics[width=\linewidth]{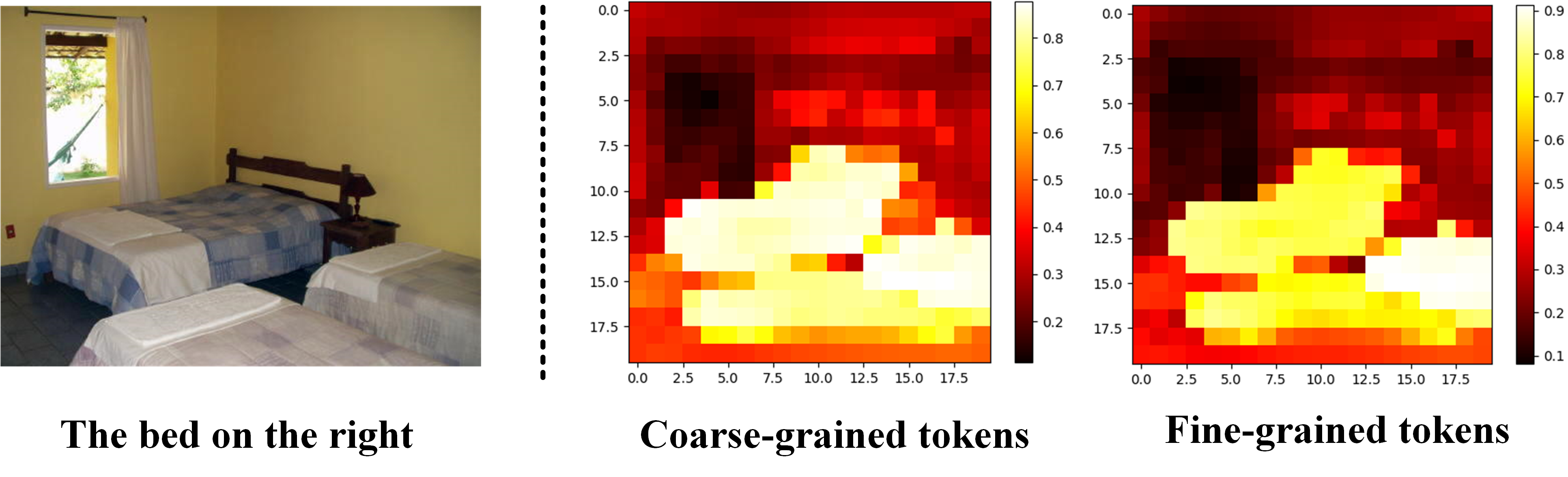} 
	\caption{Visualization of coarse-grained tokens and fine-grained tokens. Brighter pixels represent higher values, indicating that these areas receive more focus from the network.}
	\label{Fig11}
\end{figure}

\subsection{ Further Discussion}
In the proposed method, the number of decoupling phrases in the text corresponds to the number of hierarchies. On the test and validation sets of the ReferItGame and RefCOCOg datasets, we analyze the influence of the number of hierarchies on the prediction accuracy of the proposed method. Fig.~\ref{Fig9} describes the effect of the number of hierarchies on model performance. As shown in Fig.~\ref{Fig9}, as the number of hierarchies increases, the prediction accuracy of the model gradually improves. It indicates that the information extracted from each hierarchical phrase has been successfully leveraged, thereby improving the model performance.

To further understand this phenomenon, we analyze the changes in the attention maps of the CMHM at each hierarchy, as shown in Fig.~\ref{Fig10}. We find that as the hierarchy of text description increases, the attention received by the target area increases, and the attention received by the non-target area gradually decreases. It is mainly because the features at different hierarchies involve the description of the target object. Therefore, the features referring to the target object between levels are highlighted by exploiting the correlation between different hierarchies, while those related to non-target objects are suppressed. Taking ``person in middle above rock" as an example, at the first hierarchy ``person", the model pays attention to the areas where people is located. The text ``person" in this hierarchy contains multiple target locations including the target that the complete text refers to. On the second hierarchy "person in middle", since both this hierarchy and the previous hierarchy include the target person, in exploiting hierarchical relationships, the ``person" related to the target object is highlighted, and target-irrelevant ``person'' is suppressed. The third hierarchy is "person in middle above rock". Since three hierarchies involve the person and two levels involve ``person in middle'', the features related to ``person'' and ``middle'' are highlighted by the proposed method, and the person corresponding to the middle and rock receives more attention from the network.  Therefore, the proposed hierarchical cross-modal matching method can highlight the features of the target object and suppress the features of non-target objects.

To further validate the effectiveness of the hierarchical cross-modal matching method, we conduct visualizations of the coarse-grained and fine-grained tokens, as shown in Fig.~\ref{Fig11}. It is evident from these visualizations that coarse-grained tokens mainly focus on the rough visual regions corresponding to the text query. After processing through the hierarchical cross-modal matching, the fine-grained tokens are able to focus more precisely on the specific regions specified by text query. This further demonstrates the effectiveness of the proposed hierarchical cross-modal matching method in highlighting the features of the target object and suppressing the features of non-target objects.

\section{Conclusion}
In this paper, we propose a phrase decoupling cross-modal hierarchical matching and progressive position correction method for visual grounding. The method can highlight the referred features of the target object in the text by exploiting the shared information at different hierarchies. Furthermore, to effectively use these features for target localization, we develop a text and image hierarchical matching mechanism, with which a corresponding target object position progressive correction strategy is designed. It allows the position of the target object bounding box to be continuously optimized and adjusted by predicting the corrected value of the target object position. Experimental results show that the proposed method can improve the performance in terms of precision for visual grounding. The ablation experiments demonstrate the effectiveness of different components of the proposed method.

\bibliography{mybibfile}

\begin{thebibliography}{10}
\providecommand{\url}[1]{#1}
\csname url@samestyle\endcsname
\providecommand{\newblock}{\relax}
\providecommand{\bibinfo}[2]{#2}
\providecommand{\BIBentrySTDinterwordspacing}{\spaceskip=0pt\relax}
\providecommand{\BIBentryALTinterwordstretchfactor}{4}
\providecommand{\BIBentryALTinterwordspacing}{\spaceskip=\fontdimen2\font plus
\BIBentryALTinterwordstretchfactor\fontdimen3\font minus
  \fontdimen4\font\relax}
\providecommand{\BIBforeignlanguage}[2]{{%
\expandafter\ifx\csname l@#1\endcsname\relax
\typeout{** WARNING: IEEEtran.bst: No hyphenation pattern has been}%
\typeout{** loaded for the language `#1'. Using the pattern for}%
\typeout{** the default language instead.}%
\else
\language=\csname l@#1\endcsname
\fi
#2}}
\providecommand{\BIBdecl}{\relax}
\BIBdecl

\bibitem{deng2023transvg++}
J.~Deng, Z.~Yang, D.~Liu, T.~Chen, W.~Zhou, Y.~Zhang, H.~Li, and W.~Ouyang,
  ``Transvg++: End-to-end visual grounding with language conditioned vision
  transformer,'' \emph{IEEE Transactions on Pattern Analysis and Machine
  Intelligence}, 2023.

\bibitem{du2022visual}
Y.~Du, Z.~Fu, Q.~Liu, and Y.~Wang, ``Visual grounding with transformers,'' in
  \emph{2022 IEEE International Conference on Multimedia and Expo (ICME)},
  2022, pp. 1--6.

\bibitem{yang2022improving}
L.~Yang, Y.~Xu, C.~Yuan, W.~Liu, B.~Li, and W.~Hu, ``Improving visual grounding
  with visual-linguistic verification and iterative reasoning,'' in \emph{2022
  IEEE/CVF Conference on Computer Vision and Pattern Recognition (CVPR)}, 2022,
  pp. 9489--9498.

\bibitem{jiao2023suspected}
Y.~Jiao, Z.~Jie, J.~Chen, L.~Ma, and Y.-G. Jiang, ``Suspected objects matter:
  Rethinking model's prediction for one-stage visual grounding,'' in
  \emph{Proceedings of the 31st ACM International Conference on Multimedia},
  2023, pp. 17--26.

\bibitem{9470913}
Y.~Zhou, R.~Ji, G.~Luo, X.~Sun, J.~Su, X.~Ding, C.-W. Lin, and Q.~Tian, ``A
  real-time global inference network for one-stage referring expression
  comprehension,'' \emph{IEEE Transactions on Neural Networks and Learning
  Systems}, vol.~34, no.~1, pp. 134--143, 2023.

\bibitem{yang2023improving}
Z.~Yang, K.~Kafle, F.~Dernoncourt, and V.~Ordonez, ``Improving visual grounding
  by encouraging consistent gradient-based explanations,'' in \emph{2023
  IEEE/CVF Conference on Computer Vision and Pattern Recognition (CVPR)}, 2023,
  pp. 19\,165--19\,174.

\bibitem{9939075}
Y.~Bu, L.~Li, J.~Xie, Q.~Liu, Y.~Cai, Q.~Huang, and Q.~Li, ``Scene-text
  oriented referring expression comprehension,'' \emph{IEEE Transactions on
  Multimedia}, vol.~25, pp. 7208--7221, 2023.

\bibitem{wang2021exploring}
W.~Wen, C.~Yang, Z.~Jing, H.~Fengxiang, Z.~Zheng-Jun, W.~Yonggang, and
  T.~Dacheng, ``Exploring sequence feature alignment for domain adaptive
  detection transformers,'' in \emph{Proceedings of the 29th ACM International
  Conference on Multimedia}, 2021, p. 1730–1738.

\bibitem{8740604}
C.~Liu, Y.~Tao, J.~Liang, K.~Li, and Y.~Chen, ``Object detection based on yolo
  network,'' in \emph{2018 IEEE 4th Information Technology and Mechatronics
  Engineering Conference (ITOEC)}, 2018, pp. 799--803.

\bibitem{li2022comprehending}
Y.~Li, Y.~Pan, T.~Yao, and T.~Mei, ``Comprehending and ordering semantics for
  image captioning,'' in \emph{2022 IEEE/CVF Conference on Computer Vision and
  Pattern Recognition (CVPR)}, 2022, pp. 17\,969--17\,978.

\bibitem{hu2022scaling}
X.~Hu, Z.~Gan, J.~Wang, Z.~Yang, Z.~Liu, Y.~Lu, and L.~Wang, ``Scaling up
  vision-language pretraining for image captioning,'' in \emph{2022 IEEE/CVF
  Conference on Computer Vision and Pattern Recognition (CVPR)}, 2022, pp.
  17\,959--17\,968.

\bibitem{barraco2022unreasonable}
M.~Barraco, M.~Cornia, S.~Cascianelli, L.~Baraldi, and R.~Cucchiara, ``The
  unreasonable effectiveness of clip features for image captioning: An
  experimental analysis,'' in \emph{2022 IEEE/CVF Conference on Computer Vision
  and Pattern Recognition Workshops (CVPRW)}, 2022, pp. 4661--4669.

\bibitem{gupta2022swapmix}
V.~Gupta, Z.~Li, A.~Kortylewski, C.~Zhang, Y.~Li, and A.~Yuille, ``Swapmix:
  Diagnosing and regularizing the over-reliance on visual context in visual
  question answering,'' in \emph{2022 IEEE/CVF Conference on Computer Vision
  and Pattern Recognition (CVPR)}, 2022, pp. 5068--5078.

\bibitem{ding2022mukea}
Y.~Ding, J.~Yu, B.~Liu, Y.~Hu, M.~Cui, and Q.~Wu, ``Mukea: Multimodal knowledge
  extraction and accumulation for knowledge-based visual question answering,''
  in \emph{2022 IEEE/CVF Conference on Computer Vision and Pattern Recognition
  (CVPR)}, 2022, pp. 5079--5088.

\bibitem{shao2023prompting}
Z.~Shao, Z.~Yu, M.~Wang, and J.~Yu, ``Prompting large language models with
  answer heuristics for knowledge-based visual question answering,'' in
  \emph{2023 IEEE/CVF Conference on Computer Vision and Pattern Recognition
  (CVPR)}, 2023, pp. 14\,974--14\,983.

\bibitem{zhou2022towards}
Y.~Zhou, R.~Zhang, C.~Chen, C.~Li, C.~Tensmeyer, T.~Yu, J.~Gu, J.~Xu, and
  T.~Sun, ``Towards language-free training for text-to-image generation,'' in
  \emph{2022 IEEE/CVF Conference on Computer Vision and Pattern Recognition
  (CVPR)}, 2022, pp. 17\,886--17\,896.

\bibitem{li2023gligen}
Y.~Li, H.~Liu, Q.~Wu, F.~Mu, J.~Yang, J.~Gao, C.~Li, and Y.~J. Lee, ``Gligen:
  Open-set grounded text-to-image generation,'' in \emph{2023 IEEE/CVF
  Conference on Computer Vision and Pattern Recognition (CVPR)}, 2023, pp.
  22\,511--22\,521.

\bibitem{kim2023dense}
Y.~Kim, J.~Lee, J.-H. Kim, J.-W. Ha, and J.-Y. Zhu, ``Dense text-to-image
  generation with attention modulation,'' in \emph{2023 IEEE/CVF Conference on
  Computer Vision and Pattern Recognition (CVPR)}, 2023, pp. 7701--7711.

\bibitem{Alpher05}
L.~Yu, Z.~Lin, X.~Shen, J.~Yang, X.~Lu, M.~Bansal, and T.~L. Berg, ``Mattnet:
  Modular attention network for referring expression comprehension,'' in
  \emph{2018 IEEE/CVF Conference on Computer Vision and Pattern Recognition},
  2018, pp. 1307--1315.

\bibitem{8691415}
R.~Hong, D.~Liu, X.~Mo, X.~He, and H.~Zhang, ``Learning to compose and reason
  with language tree structures for visual grounding,'' \emph{IEEE Transactions
  on Pattern Analysis and Machine Intelligence}, vol.~44, no.~2, pp. 684--696,
  2022.

\bibitem{9009000}
D.~Liu, H.~Zhang, Z.-J. Zha, and F.~Wu, ``Learning to assemble neural module
  tree networks for visual grounding,'' in \emph{2019 IEEE/CVF International
  Conference on Computer Vision (ICCV)}, 2019, pp. 4672--4681.

\bibitem{chen2021ref}
L.~Chen, W.~Ma, J.~Xiao, H.~Zhang, and S.-F. Chang, ``Ref-nms: Breaking
  proposal bottlenecks in two-stage referring expression grounding,'' in
  \emph{Proceedings of the AAAI Conference on Artificial Intelligence}, 2021,
  pp. 1036--1044.

\bibitem{9806393}
H.~Zhao, J.~T. Zhou, and Y.-S. Ong, ``Word2pix: Word to pixel cross-attention
  transformer in visual grounding,'' \emph{IEEE Transactions on Neural Networks
  and Learning Systems}, pp. 1--11, 2022.

\bibitem{9699024}
M.~Sun, W.~Suo, P.~Wang, Y.~Zhang, and Q.~Wu, ``A proposal-free one-stage
  framework for referring expression comprehension and generation via dense
  cross-attention,'' \emph{IEEE Transactions on Multimedia}, vol.~25, pp.
  2446--2458, 2023.

\bibitem{10035452}
G.~Hua, M.~Liao, S.~Tian, Y.~Zhang, and W.~Zou, ``Multiple relational learning
  network for joint referring expression comprehension and segmentation,''
  \emph{IEEE Transactions on Multimedia}, pp. 1--13, 2023.

\bibitem{9982652}
W.~Suo, M.~Sun, P.~Wang, Y.~Zhang, and Q.~Wu, ``Rethinking and improving
  feature pyramids for one-stage referring expression comprehension,''
  \emph{IEEE Transactions on Image Processing}, vol.~32, pp. 854--864, 2023.

\bibitem{9710016}
J.~Deng, Z.~Yang, T.~Chen, W.~Zhou, and H.~Li, ``Transvg: End-to-end visual
  grounding with transformers,'' in \emph{2021 IEEE/CVF International
  Conference on Computer Vision (ICCV)}, 2021, pp. 1749--1759.

\bibitem{huang2021look}
B.~Huang, D.~Lian, W.~Luo, and S.~Gao, ``Look before you leap: Learning
  landmark features for one-stage visual grounding,'' in \emph{2021 IEEE/CVF
  Conference on Computer Vision and Pattern Recognition (CVPR)}, 2021, pp.
  16\,883--16\,892.

\bibitem{qiu2020language}
H.~Qiu, H.~Li, Q.~Wu, F.~Meng, H.~Shi, T.~Zhao, and K.~N. Ngan,
  ``Language-aware fine-grained object representation for referring expression
  comprehension,'' in \emph{Proceedings of the 28th ACM International
  Conference on Multimedia}, 2020, pp. 4171--4180.

\bibitem{ye2021one}
J.~Ye, X.~Lin, L.~He, D.~Li, and Q.~Chen, ``One-stage visual grounding via
  semantic-aware feature filter,'' in \emph{Proceedings of the 29th ACM
  International Conference on Multimedia}, 2021, pp. 1702--1711.

\bibitem{girshick2015fast}
R.~Girshick, ``Fast r-cnn,'' in \emph{2015 IEEE International Conference on
  Computer Vision (ICCV)}, 2015, pp. 1440--1448.

\bibitem{he2017mask}
K.~He, G.~Gkioxari, P.~Dollár, and R.~Girshick, ``Mask r-cnn,'' in \emph{2017
  IEEE International Conference on Computer Vision (ICCV)}, 2017, pp.
  2980--2988.

\bibitem{jing2020visual}
C.~Jing, Y.~Wu, M.~Pei, Y.~Hu, Y.~Jia, and Q.~Wu, ``Visual-semantic graph
  matching for visual grounding,'' in \emph{Proceedings of the 28th ACM
  International Conference on Multimedia}, 2020, pp. 4041--4050.

\bibitem{yang2019dynamic}
S.~Yang, Y.~Yu, and G.~Li, ``Dynamic graph attention for referring expression
  comprehension,'' in \emph{2019 IEEE/CVF International Conference on Computer
  Vision (ICCV)}, 2019, pp. 4643--4652.

\bibitem{yang2019cross}
S.~Yang, G.~Li, and Y.~Yu, ``Cross-modal relationship inference for grounding
  referring expressions,'' in \emph{2019 IEEE/CVF Conference on Computer Vision
  and Pattern Recognition (CVPR)}, 2019, pp. 4140--4149.

\bibitem{8999516}
S.~Yang, Y.~Yu, and G.~Li, ``Relationship-embedded representation learning for
  grounding referring expressions,'' \emph{IEEE Transactions on Pattern
  Analysis and Machine Intelligence}, vol.~43, no.~8, pp. 2765--2779, 2021.

\bibitem{liu2019improving}
X.~Liu, Z.~Wang, J.~Shao, X.~Wang, and H.~Li, ``Improving referring expression
  grounding with cross-modal attention-guided erasing,'' in \emph{2019 IEEE/CVF
  Conference on Computer Vision and Pattern Recognition (CVPR)}, 2019, pp.
  1950--1959.

\bibitem{yang2019fast}
Z.~Yang, B.~Gong, L.~Wang, W.~Huang, D.~Yu, and J.~Luo, ``A fast and accurate
  one-stage approach to visual grounding,'' in \emph{2019 IEEE/CVF
  International Conference on Computer Vision (ICCV)}, 2019, pp. 4682--4692.

\bibitem{redmon2018yolov3}
J.~Redmon and A.~Farhadi, ``Yolov3: An incremental improvement,'' \emph{arXiv
  preprint arXiv:1804.02767}, 2018.

\bibitem{yang2020improving}
Z.~Yang, T.~Chen, L.~Wang, and J.~Luo, ``Improving one-stage visual grounding
  by recursive sub-query construction,'' in \emph{Computer Vision -- ECCV
  2020}, 2020, pp. 387--404.

\bibitem{liao2020real}
Y.~Liao, S.~Liu, G.~Li, F.~Wang, Y.~Chen, C.~Qian, and B.~Li, ``A real-time
  cross-modality correlation filtering method for referring expression
  comprehension,'' in \emph{2020 IEEE/CVF Conference on Computer Vision and
  Pattern Recognition (CVPR)}, 2020, pp. 10\,877--10\,886.

\bibitem{vaswani2017attention}
A.~Vaswani, N.~Shazeer, N.~Parmar, J.~Uszkoreit, L.~Jones, A.~N. Gomez,
  {\L}.~Kaiser, and I.~Polosukhin, ``Attention is all you need,''
  \emph{Advances in neural information processing systems}, vol.~30, 2017.

\bibitem{chen2021crossvit}
C.-F.~R. Chen, Q.~Fan, and R.~Panda, ``Crossvit: Cross-attention multi-scale
  vision transformer for image classification,'' in \emph{2021 IEEE/CVF
  Conference on Computer Vision and Pattern Recognition (CVPR)}, 2021, pp.
  357--366.

\bibitem{dosovitskiy2020image}
A.~Dosovitskiy, L.~Beyer, A.~Kolesnikov, D.~Weissenborn, X.~Zhai,
  T.~Unterthiner, M.~Dehghani, M.~Minderer, G.~Heigold, S.~Gelly \emph{et~al.},
  ``An image is worth 16x16 words: Transformers for image recognition at
  scale,'' \emph{arXiv preprint arXiv:2010.11929}, 2020.

\bibitem{10066256}
B.~Xu, X.~Shu, J.~Zhang, G.~Dai, and Y.~Song, ``Spatiotemporal
  decouple-and-squeeze contrastive learning for semisupervised skeleton-based
  action recognition,'' \emph{IEEE Transactions on Neural Networks and Learning
  Systems}, pp. 1--14, 2023.

\bibitem{xu2023pyramid}
B.~Xu and X.~Shu, ``Pyramid self-attention polymerization learning for
  semi-supervised skeleton-based action recognition,'' \emph{arXiv preprint
  arXiv:2302.02327}, 2023.

\bibitem{9954217}
X.~Shu, B.~Xu, L.~Zhang, and J.~Tang, ``Multi-granularity anchor-contrastive
  representation learning for semi-supervised skeleton-based action
  recognition,'' \emph{IEEE Transactions on Pattern Analysis and Machine
  Intelligence}, vol.~45, no.~6, pp. 7559--7576, 2023.

\bibitem{9782720}
B.~Xu, X.~Shu, and Y.~Song, ``X-invariant contrastive augmentation and
  representation learning for semi-supervised skeleton-based action
  recognition,'' \emph{IEEE Transactions on Image Processing}, vol.~31, pp.
  3852--3867, 2022.

\bibitem{ye2022shifting}
J.~Ye, J.~Tian, M.~Yan, X.~Yang, X.~Wang, J.~Zhang, L.~He, and X.~Lin,
  ``Shifting more attention to visual backbone: Query-modulated refinement
  networks for end-to-end visual grounding,'' in \emph{2022 IEEE/CVF Conference
  on Computer Vision and Pattern Recognition (CVPR)}, 2022, pp.
  15\,481--15\,491.

\bibitem{zhan2023rsvg}
Y.~Zhan, Z.~Xiong, and Y.~Yuan, ``Rsvg: Exploring data and models for visual
  grounding on remote sensing data,'' \emph{IEEE Transactions on Geoscience and
  Remote Sensing}, vol.~61, pp. 1--13, 2023.

\bibitem{10298801}
F.~Shi, R.~Gao, W.~Huang, and L.~Wang, ``Dynamic mdetr: A dynamic multimodal
  transformer decoder for visual grounding,'' \emph{IEEE Transactions on
  Pattern Analysis and Machine Intelligence}, pp. 1--18, 2023.

\bibitem{10269126}
L.~Xiao, X.~Yang, F.~Peng, M.~Yan, Y.~Wang, and C.~Xu, ``Clip-vg: Self-paced
  curriculum adapting of clip for visual grounding,'' \emph{IEEE Transactions
  on Multimedia}, pp. 1--14, 2023.

\bibitem{radford2021learning}
A.~Radford, J.~W. Kim, C.~Hallacy, A.~Ramesh, G.~Goh, S.~Agarwal, G.~Sastry,
  A.~Askell, P.~Mishkin, J.~Clark \emph{et~al.}, ``Learning transferable visual
  models from natural language supervision,'' in \emph{Proceedings of the 38th
  International Conference on Machine Learning}, 2021, pp. 8748--8763.

\bibitem{akbik2018coling}
A.~Akbik, D.~Blythe, and R.~Vollgraf, ``Contextual string embeddings for
  sequence labeling,'' in \emph{Proceedings of the 27th International
  Conference on Computational Linguistics}, 2018, pp. 1638--1649.

\bibitem{kenton2019bert}
J.~D. M.-W.~C. Kenton and L.~K. Toutanova, ``Bert: Pre-training of deep
  bidirectional transformers for language understanding,'' in \emph{Proceedings
  of naacL-HLT}, vol.~1, 2019, p.~2.

\bibitem{8099953}
R.~Hu, M.~Rohrbach, J.~Andreas, T.~Darrell, and K.~Saenko, ``Modeling
  relationships in referential expressions with compositional modular
  networks,'' in \emph{2017 IEEE Conference on Computer Vision and Pattern
  Recognition (CVPR)}, 2017, pp. 4418--4427.

\bibitem{8753569}
Y.~Niu, H.~Zhang, Z.~Lu, and S.-F. Chang, ``Variational context: Exploiting
  visual and textual context for grounding referring expressions,'' \emph{IEEE
  Transactions on Pattern Analysis and Machine Intelligence}, vol.~43, no.~1,
  pp. 347--359, 2021.

\bibitem{8578545}
B.~Zhuang, Q.~Wu, C.~Shen, I.~Reid, and A.~v.~d. Hengel, ``Parallel attention:
  A unified framework for visual object discovery through dialogs and
  queries,'' in \emph{2018 IEEE/CVF Conference on Computer Vision and Pattern
  Recognition}, 2018, pp. 4252--4261.

\bibitem{9157418}
G.~Luo, Y.~Zhou, X.~Sun, L.~Cao, C.~Wu, C.~Deng, and R.~Ji, ``Multi-task
  collaborative network for joint referring expression comprehension and
  segmentation,'' in \emph{2020 IEEE/CVF Conference on Computer Vision and
  Pattern Recognition (CVPR)}, 2020, pp. 10\,031--10\,040.

\bibitem{9578259}
B.~Huang, D.~Lian, W.~Luo, and S.~Gao, ``Look before you leap: Learning
  landmark features for one-stage visual grounding,'' in \emph{2021 IEEE/CVF
  Conference on Computer Vision and Pattern Recognition (CVPR)}, 2021, pp.
  16\,883--16\,892.

\bibitem{zhu2022seqtr}
C.~Zhu, Y.~Zhou, Y.~Shen, G.~Luo, X.~Pan, M.~Lin, C.~Chen, L.~Cao, X.~Sun, and
  R.~Ji, ``Seqtr: A simple yet universal network for visual grounding,'' in
  \emph{European Conference on Computer Vision}, 2022, pp. 598--615.

\bibitem{9711011}
Y.~Zhou, T.~Ren, C.~Zhu, X.~Sun, J.~Liu, X.~Ding, M.~Xu, and R.~Ji, ``Trar:
  Routing the attention spans in transformer for visual question answering,''
  in \emph{2021 IEEE/CVF International Conference on Computer Vision (ICCV)},
  2021, pp. 2054--2064.

\bibitem{wu2024improving}
J.~Wu, C.~Wu, F.~Wang, L.~Wang, and Y.~Wei, ``Improving visual grounding with
  multi-scale discrepancy information and centralized-transformer,''
  \emph{Expert Systems with Applications}, vol. 247, p. 123223, 2024.

\bibitem{liao2022progressive}
Y.~Liao, A.~Zhang, Z.~Chen, T.~Hui, and S.~Liu, ``Progressive
  language-customized visual feature learning for one-stage visual grounding,''
  \emph{IEEE Transactions on Image Processing}, vol.~31, pp. 4266--4277, 2022.

\bibitem{liang2023luna}
Y.~Liang, Z.~Yang, Y.~Tang, J.~Fan, Z.~Li, J.~Wang, P.~H. Torr, and S.-L.
  Huang, ``Luna: Language as continuing anchors for referring expression
  comprehension,'' in \emph{Proceedings of the 31th ACM International
  Conference on Multimedia}, 2023, pp. 5174--5184.

\bibitem{10285487}
H.~Zhu, Q.~Lu, L.~Xue, M.~Xue, G.~Yuan, and B.~Zhong, ``Visual grounding with
  joint multimodal representation and interaction,'' \emph{IEEE Transactions on
  Instrumentation and Measurement}, vol.~72, pp. 1--11, 2023.

\bibitem{8953982}
H.~Rezatofighi, N.~Tsoi, J.~Gwak, A.~Sadeghian, I.~Reid, and S.~Savarese,
  ``Generalized intersection over union: A metric and a loss for bounding box
  regression,'' in \emph{2019 IEEE/CVF Conference on Computer Vision and
  Pattern Recognition (CVPR)}, 2019, pp. 658--666.

\bibitem{1111111}
L.~Yu, P.~Poirson, S.~Yang, A.~C. Berg, and T.~L. Berg, ``Modeling context in
  referring expressions,'' in \emph{Computer Vision--ECCV 2016: 14th European
  Conference, Amsterdam, The Netherlands, October 11-14, 2016, Proceedings,
  Part II 14}, 2016, pp. 69--85.

\bibitem{7780378}
J.~Mao, J.~Huang, A.~Toshev, O.~Camburu, A.~Yuille, and K.~Murphy, ``Generation
  and comprehension of unambiguous object descriptions,'' in \emph{2016 IEEE
  Conference on Computer Vision and Pattern Recognition (CVPR)}, 2016, pp.
  11--20.

\bibitem{kazemzadeh2014referitgame}
S.~Kazemzadeh, V.~Ordonez, M.~Matten, and T.~Berg, ``Referitgame: Referring to
  objects in photographs of natural scenes,'' in \emph{Proceedings of the 2014
  conference on empirical methods in natural language processing (EMNLP)},
  2014, pp. 787--798.

\bibitem{vinyals2016show}
O.~Vinyals, A.~Toshev, S.~Bengio, and D.~Erhan, ``Show and tell: Lessons
  learned from the 2015 mscoco image captioning challenge,'' \emph{IEEE
  transactions on pattern analysis and machine intelligence}, vol.~39, no.~4,
  pp. 652--663, 2016.

\bibitem{escalante2010segmented}
H.~J. Escalante, C.~A. Hern{\'a}ndez, J.~A. Gonzalez, A.~L{\'o}pez-L{\'o}pez,
  M.~Montes, E.~F. Morales, L.~E. Sucar, L.~Villasenor, and M.~Grubinger, ``The
  segmented and annotated iapr tc-12 benchmark,'' \emph{Computer vision and
  image understanding}, vol. 114, no.~4, pp. 419--428, 2010.

\bibitem{he2016deep}
K.~He, X.~Zhang, S.~Ren, and J.~Sun, ``Deep residual learning for image
  recognition,'' in \emph{Proceedings of the IEEE conference on computer vision
  and pattern recognition}, 2016, pp. 770--778.

\bibitem{loshchilov2018decoupled}
I.~Loshchilov and F.~Hutter, ``Decoupled weight decay regularization,'' in
  \emph{International Conference on Learning Representations}, 2018.

\bibitem{liu2021swin}
Z.~Liu, Y.~Lin, Y.~Cao, H.~Hu, Y.~Wei, Z.~Zhang, S.~Lin, and B.~Guo, ``Swin
  transformer: Hierarchical vision transformer using shifted windows,'' in
  \emph{Proceedings of the IEEE/CVF international conference on computer
  vision}, 2021, pp. 10\,012--10\,022.

\end{thebibliography}
\end{document}